%% file: aistats2024_spurious.tex
\documentclass[twoside]{article}

\usepackage[accepted]{aistats2024}

\usepackage[round]{natbib}

\usepackage{url}

\usepackage{microtype}
\usepackage{graphicx}
\usepackage{subcaption}
\usepackage{booktabs} %
\usepackage{xspace}
\usepackage[normalem]{ulem}
\usepackage{hyperref}
\usepackage{multirow}

\usepackage{algorithm}
\usepackage{algorithmic}

\renewcommand{\algorithmicrequire}{\textbf{Input:}}
\renewcommand{\algorithmicensure}{\textbf{Output:}}

\usepackage[utf8]{inputenc} %
\usepackage[T1]{fontenc}    %
\usepackage{hyperref}       %
\usepackage{url}            %
\usepackage{booktabs}       %
\usepackage{amsfonts}       %
\usepackage{nicefrac}       %
\usepackage{microtype}      %
\usepackage{xcolor}         %

\usepackage{amsmath}
\usepackage{amssymb}
\usepackage{mathtools}
\usepackage{amsthm}
\usepackage{bbm}
\usepackage{bm}
\usepackage{enumitem}
\usepackage{comment}
\usepackage{xcolor,colortbl}
\usepackage{thmtools}
\usepackage{thm-restate}

\usepackage[toc,page,titletoc]{appendix}
\usepackage[capitalise,nameinlink]{cleveref}
\crefname{section}{Sec.}{Sec.}
\crefname{appendix}{Sec.}{Sec.}
\crefname{table}{Tab.}{Tab.}
\crefname{algorithm}{Alg.}{Alg.}

\usepackage[export]{adjustbox}
\usepackage{array}
\usepackage{tabularx}
\usepackage{booktabs}
\usepackage{graphicx}
\usepackage{wrapfig,lipsum,booktabs}
\usepackage{arydshln}

\usepackage[varbb]{newtxmath}

\theoremstyle{plain}
\newtheorem{theorem}{Theorem}[section]

\theoremstyle{definition}

\newtheorem{assumption}[theorem]{Assumption}
\theoremstyle{remark}

\usepackage[textsize=tiny]{todonotes}

\newcommand{\x}{\pmb{x}}
\newcommand{\X}{\pmb{X}}

\newcommand{\w}{\pmb{w}}

\newcommand{\vb}{\pmb{v}}
\newcommand{\z}{\pmb{z}}
\newcommand{\D}{\mathcal{D}}
\newcommand{\OO}{\mathcal{O}}
\newcommand{\LL}{\mathcal{L}}

\newcommand{\A}{\mathcal{A}}

\newcommand{\R}{\mathbb{R}}
\newcommand{\W}{\pmb{W}}

\newcommand{\bet}{\pmb{\beta}}
\newcommand{\mtx}{\bm} %
\newcommand{\vct}{\bm} %
\newcommand{\mino}{\text{mino}}
\newcommand{\maj}{\text{maj}}
\newcommand{\inner}[2]{\left\langle#1,#2\right\rangle}

\newcommand{\norm}[1]{ \left\| #1 \right\| }
\DeclareMathOperator{\Tr}{Tr}

\DeclareMathOperator*{\E}{\mathbb{E}}

\newcommand{\alg}{\textsc{Spare}\xspace} 
\newcommand{\fullname}{\textsc{S}e\textsc{PA}rate early and \textsc{RE}sample\xspace} 

\begin{document}
\runningtitle{\alg: Identifying Spurious Biases Early in Training}

\runningauthor{Yu Yang, Eric Gan, Gintare Karolina Dziugaite, Baharan Mirzasoleiman}

\twocolumn[

\aistatstitle{Identifying Spurious Biases Early in Training \\ through the Lens of Simplicity Bias}

\aistatsauthor{ Yu Yang \And Eric Gan}
\aistatsaddress{ University of California, Los Angeles \And University of California, Los Angeles }
\aistatsauthor{Gintare Karolina Dziugaite \And Baharan Mirzasoleiman }
\aistatsaddress{ Google Deepmind \And University of California, Los Angeles} ]

\begin{abstract}
\input{sections/0_abstract}

\end{abstract}

\input{sections/1_intro_2}
\input{sections/2_related_2}

\input{sections/3_method}

\input{sections/4_experiments}
\input{sections/5_conclusion}

\subsubsection*{Acknowledgements}
This research was partially supported by the National Science Foundation
CAREER Award 2146492, Amazon PhD fellowship, 
and Google Cloud research credits.

\bibliography{subpopulation}

\newpage
\section*{Checklist}

 \begin{enumerate}

 \item For all models and algorithms presented, check if you include:
 \begin{enumerate}
   \item A clear description of the mathematical setting, assumptions, algorithm, and/or model. [Yes]
   \item An analysis of the properties and complexity (time, space, sample size) of any algorithm. [Yes]
   \item (Optional) Anonymized source code, with specification of all dependencies, including external libraries. [Yes]
 \end{enumerate}

 \item For any theoretical claim, check if you include:
 \begin{enumerate}
   \item Statements of the full set of assumptions of all theoretical results. [Yes]
   \item Complete proofs of all theoretical results. [Yes]
   \item Clear explanations of any assumptions. [Yes]     
 \end{enumerate}

 \item For all figures and tables that present empirical results, check if you include:
 \begin{enumerate}
   \item The code, data, and instructions needed to reproduce the main experimental results (either in the supplemental material or as a URL). [Yes]
   \item All the training details (e.g., data splits, hyperparameters, how they were chosen). [Yes]
         \item A clear definition of the specific measure or statistics and error bars (e.g., with respect to the random seed after running experiments multiple times). [Yes]
         \item A description of the computing infrastructure used. (e.g., type of GPUs, internal cluster, or cloud provider). [Yes]
 \end{enumerate}

 \item If you are using existing assets (e.g., code, data, models) or curating/releasing new assets, check if you include:
 \begin{enumerate}
   \item Citations of the creator If your work uses existing assets. [Yes]
   \item The license information of the assets, if applicable. [Yes]
   \item New assets either in the supplemental material or as a URL, if applicable. [Not Applicable]
   \item Information about consent from data providers/curators. [Not Applicable]
   \item Discussion of sensible content if applicable, e.g., personally identifiable information or offensive content. [Not Applicable]
 \end{enumerate}

 \item If you used crowdsourcing or conducted research with human subjects, check if you include:
 \begin{enumerate}
   \item The full text of instructions given to participants and screenshots. [Not Applicable]
   \item Descriptions of potential participant risks, with links to Institutional Review Board (IRB) approvals if applicable. [Not Applicable]
   \item The estimated hourly wage paid to participants and the total amount spent on participant compensation. [Not Applicable]
 \end{enumerate}

 \end{enumerate}

\newpage
\onecolumn
\appendix

\input{sections/6_appendix}

\end{document}

%% file: sections/0_abstract.tex
Neural networks trained with (stochastic) gradient descent have an inductive bias towards learning simpler solutions. %
This makes them highly prone to learning \textit{spurious correlations} in the training data, that may not hold at test time. %
In this work, we provide the first theoretical analysis of the effect of simplicity bias on learning spurious correlations. 
Notably, we show that
examples with spurious features
are \textit{provably} separable based on the model's output \textit{early in training}. We further illustrate that if spurious features have a small enough {noise-to-signal ratio}, the network’s output on majority of examples
is almost exclusively determined by the spurious features, %
leading to poor \textit{worst-group} test accuracy.
Finally, we propose \alg, which identifies spurious correlations
early in training, and utilizes importance sampling to alleviate their effect. %
Empirically, we 
demonstrate that \alg outperforms %
state-of-the-art methods by up to
21.1\% in worst-group accuracy, %
while being up to 12x faster. 
We also show that \alg is a highly effective but lightweight method to \textit{discover spurious correlations}.
Code is available at \href{https://github.com/BigML-CS-UCLA/SPARE}{https://github.com/BigML-CS-UCLA/SPARE}.\looseness=-1

%% file: sections/1_intro_2.tex
\section{INTRODUCTION}
The \textit{simplicity bias} of gradient-based training algorithms towards learning simpler solutions has been suggested as a key factor for the superior generalization performance of overparameterized neural networks \citep{hermann2020shapes,hu2020surprising,nakkiran2019sgd,neyshabur2014search,pezeshki2021gradient,shah2020pitfalls}.
At the same time, it is conjectured to make neural networks vulnerable to learning \textit{spurious correlations} frequently found in real-world datasets \citep{sagawa2019distributionally,sohoni2020no}. 
Neural networks trained with gradient-based methods can exclusively rely on
simple \textit{spurious features} that are highly correlated with a class in the training data
but are not predictive of the class in general, 
and remain invariant to the predictive but more complex \textit{core features} \citep{shah2020pitfalls}. 
This results in a poor \textit{worst-group test accuracy} on groups of examples where the spurious correlations do not hold
\citep{shah2020pitfalls,teney2022evading}. For example, in an image classification task, if the majority of images of a `bird' appear on a `sky' background, the classifier learns the sky instead of bird, and misclassifies birds that do not appear in the sky at test time.

An effective way to mitigate a spurious correlation and improve the worst-group test accuracy is to upweight examples that do not contain the spurious feature during training \citep{sagawa2019distributionally}. 
However, inspecting all training examples to find such examples becomes prohibitive in real-world datasets.
This has motivated a growing body of work on group inference: separating majority groups exhibiting spurious correlation with a class, from minority groups without the spurious correlation. 
Such methods first train a neural network with gradient methods to learn the spurious correlation. 
Then, they %
rely on model's
misclassification \citep{liu2021just}, %
loss \citep{creager2021environment}, or representations \citep{sohoni2020no,ahmed2020systematic} at a certain point during training, as an indicative of minority examples. %
The time of group inference and how much to upweight the minority groups are heavily tuned based on a group-labeled validation data \citep{sohoni2020no,liu2021just,creager2021environment,ahmed2020systematic}.\looseness=-1

Despite their success on simple benchmark datasets, we show that 
such methods
suffer from several issues: 
(1) they often misidentify minority examples as majority and mistakenly downweight them; then, %
(2) %
to counteract the spurious correlation
they need to heavily upweight their small inferred minority group. This %
magnifies milder spurious correlations that may exist in the minority group \citep{li2023whac} and harms the performance;
as (3) there is no theoretical guideline for finding the time of group inference and group weights, such methods rely on extensive hyperparameter tuning. This limits their applicability and scalability.

In this work, we make several theoretical and empirical contributions towards addressing the above issues.
First, we prove that
the simplicity bias of gradient descent %
can be leveraged to identify spurious correlations. %
We analyze a two-layer fully connected neural network trained with SGD, and leverage recent results showing its early-time learning dynamics can be mimicked by training a linear model on the inputs \citep{hu2020surprising}. 
We show that %
the contribution of a spurious feature to the network output in the initial training phase increases linearly with the amount of spurious correlation. 
Thus, minority and majority groups can be 
\textit{provably} separated based on the model's output, \textit{early in training}. %
This enables more accurate identification of minorities, and limits the range of group inference to the first few training epochs, {without} extensive hyperparameter tuning.

Next, we show that once
the initial linear model converges, if the noise-to-signal ratio of a spurious feature is lower than that of the core feature in a class, the network will not learn the core features of the majority groups.
This %
explains prior empirical observations \citep{shah2020pitfalls}, by revealing \textit{when and why} neural networks trained with gradient almost exclusively rely on spurious features and remain invariant to the predictive but more
complex core features.
To the best of our knowledge, this is the first analysis of the effect of SGD's simplicity bias on learning spurious vs core features.\looseness=-1

Finally, we propose an efficient and lightweight method, \alg (\fullname), that clusters model's output early in training, %
and leverage importance sampling based on {inverse cluster sizes} 
to mitigate spurious correlations. %
This results in a superior 
worst-group accuracy on more challenging tasks, without increasing the training time, or requiring extensive hyperparameter tuning. 
Unlike existing methods, \alg can operate without a group-labeled validation data, which allows it to \textit{discover unknown spurious correlations}.\looseness=-1

Our extensive experiments confirm that \alg achieves up to 42.9\% higher worst-group accuracy over state-of-the-art on most commonly used benchmarks, including CMNIST \citep{alain2015variance} (with multiple minority groups), Waterbirds \citep{sagawa2019distributionally}, CelebA \citep{liu2015faceattributes} and UrbanCars \citep{li2023whac} (with multiple spurious correlations) while being up to 12x faster. On CMNIST, \alg performs well across varying noise-to-signal ratios, whereas other state-of-the-art methods struggle. Applied to Restricted ImageNet, a dataset {without known spurious correlations} or group-labeled validation set available for hyperparameter tuning, 
identifies the spurious correlation much more effectively
than the state-of-the-art group inference methods and improves the model's accuracy on minority groups
by up to 23.2\% higher than them after robust training.
 \looseness=-1

%% file: sections/2_related_2.tex
\section{RELATED WORK}\label{sec:related}
\paragraph{Mitigating Spurious Correlations With Group Inference.} 
To mitigate spurious correlations, group inference methods typically fall into three categories:
(1) end-of-training clustering, (2) Environment Invariance Maximization, and (3) misclassification methods.
GEORGE \citep{sohoni2020no} utilizes end-of-training clustering by first training a model using Empirical Risk Minimization (ERM). The clustered feature representations are then used to train a robust model with Group Distributionally Robust Optimization (GDRO) \citep{sagawa2019distributionally}. Methods use Environment Invariance Maximization, EIIL \citep{creager2021environment} and PGI \citep{ahmed2020systematic}, train an initial model using ERM and then partition the data to maximize the Invariant Risk Minimization (IRM) objective \citep{arjovsky2017towards}. Then, EIIL trains the robust model with GDRO, whereas PGI minimizes the KL divergence of softmaxed logits for same-class samples across groups. The misclassification approach, exemplified by JTT \citep{liu2021just} trains an ERM model for some epochs and identifies misclassified examples. The training set is then upsampled with these examples, and a robust model is trained using ERM on this augmented set.\looseness=-1

State-of-the-art methods commonly misidentify minority examples as the majority and struggle to mitigate spurious correlations. %
They heavily rely on a group-labeled validation for hyperparameter tuning,
and often increase training time during group inference or robust training. In contrast, guided by theory, \alg can %
accurately separate groups with spurious features %
in the first few epochs,
eliminating the need for extensive hyperparameter tuning and achieving better performance on minority groups without extending training time.
\looseness=-1

\paragraph{Simplicity Bias.}
Simplicity bias of SGD 
in learning simpler functions before complex ones 
\citep{hermann2020shapes,hu2020surprising,nakkiran2019sgd,neyshabur2014search,pezeshki2021gradient,shah2020pitfalls} is empirically observed in various network architectures, including MobileNetV2, ResNet50, and DenseNet121 \citep{sandler2018mobilenetv2,he2016deep,shah2020pitfalls}. \cite{hu2020surprising} formally proved that initial learning dynamics of a two-layer FC network can be mimicked by a linear model and extended this to multi-layer FC and convolutional networks. 
While simplicity bias helps overparameterized networks generalize well, it is also \textit{conjectured} to favor simpler features over complex ones, even if they are less predictive
\citep{shah2020pitfalls,teney2022evading}. However, the exact notion of the simplicity of features and the mechanism by which they are learned remain poorly understood except in certain simplistic settings \citep{nagarajan2020understanding,shah2020pitfalls}. In this work, we build on \cite{hu2020surprising} to rigorously specify the conditions and mechanism of learning spurious features in a two-layer FC network.\looseness=-1

%% file: sections/3_method.tex
\section{PROBLEM FORMULATION}
\label{sec:problem}

Let $\mathcal{D}\!=\!\{(\x_i,y_i)\}_{i=1}^n\!\subset \R^d\!\times \R$ be $n$ training data with features $\x_i\!\in\!\mathbb{R}^d$, and labels $ y_i\!\in\! \mathcal{C}\!=\!\{1,\!-1\}$. For simplicity, we consider binary classification with $\ell_2$ loss, but our analysis generalizes to multi-class classification with CE loss, and other model architectures, as we will confirm experimentally.

\begin{figure}[t]
     \centering
\includegraphics[width=0.77\columnwidth]{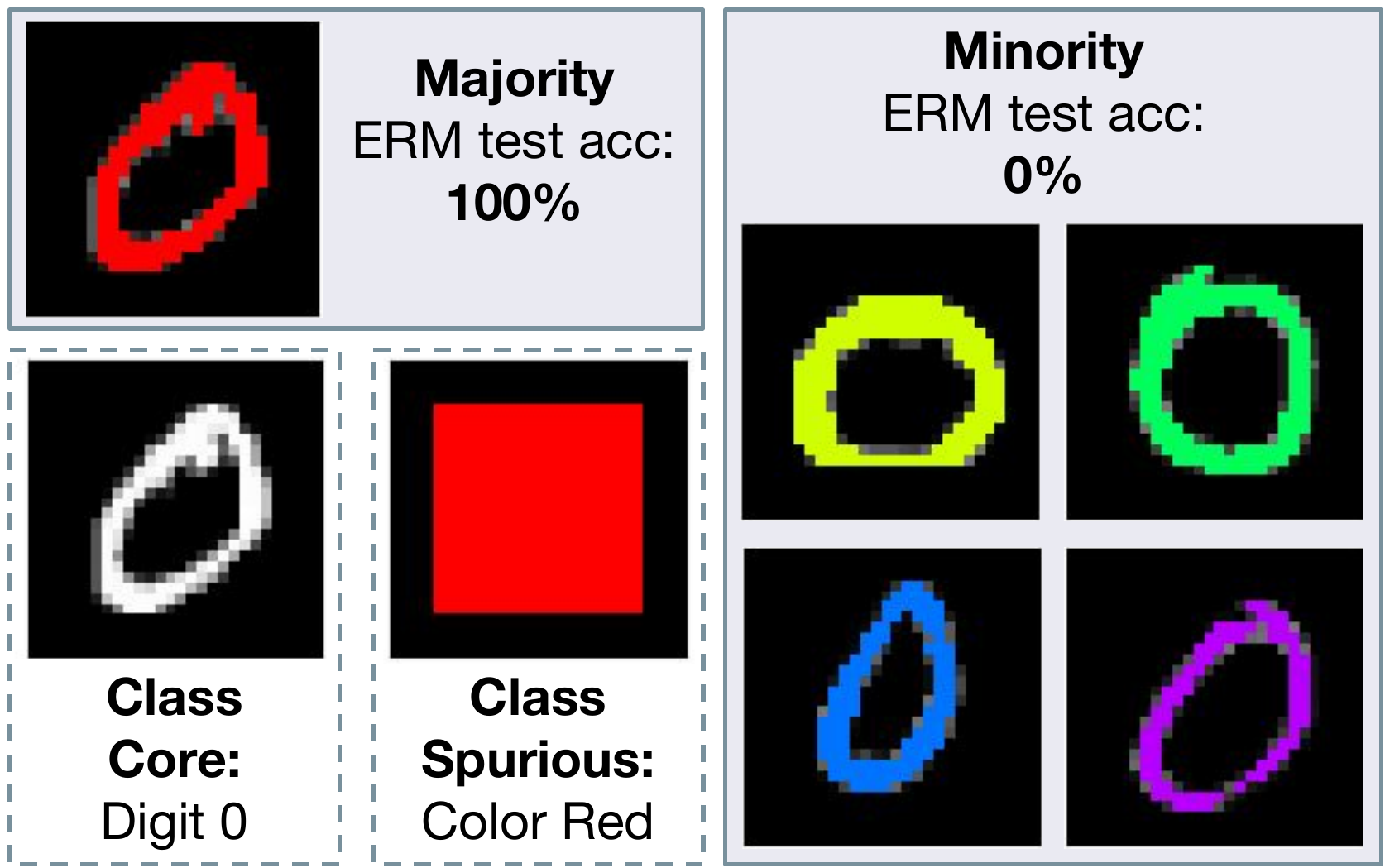}
\vspace{3mm}
     \caption{Colored MNIST as an example of datasets containing spurious correlations. Each digit is a class; the majority of digits in a class have a particular color, and the remaining digits are in other colors. Models trained with ERM learn to rely on spurious features (colors) instead of the core feature (digits) and thus do not perform well on groups of examples where the spurious correlation does not hold.}\label{fig:cmnist}
\end{figure}

\paragraph{Features \& Groups.} We assume every class $c\in\mathcal{C}$ has a \textit{core} feature $\vb_{c}$, which is the invariant feature of the class that appears in both training and test sets. Besides, there is a set of \textit{spurious} features $\A$ that are shared between classes 
and are present in both training and test sets, but may not have a spurious correlation with the labels at test time.
For example, in the CMNIST dataset containing images of colored hand-written digits (\cref{fig:cmnist}), 
the digit is the core feature, and its color is the spurious feature. 
Assuming w.l.o.g. that all $\vb_{c}, \vb_{s}\in\mathbb{R}^{d}$ are orthogonal vectors, the feature vector of every example $\x_i$ in class $c$ can be written as $\x_i=\vct{v}_{c}+\vct{v}_{s}
+\vct{\xi}_i$, 
where $\vct{v}_{s}\in\A$, %
and each $\vct{\xi}_i$ is a noise vector drawn i.i.d. from $\mathcal{N}(\vct{0}, \mtx{\Sigma}_{\xi})$. 
Noise-to-signal (NSR) ratio of a feature is defined as its variance over magnitude, i.e., $R_{.}=\sigma_{.}/\|\vct{v}_{.}\|$. Features with smaller NSR are simpler to learn.
Training examples can be partitioned into groups $g_{c, s}$ based on the combinations of their core and spurious features $(\vb_c, \vb_s)$. 
\looseness=-1

\paragraph{Neural Network \& Training.} We consider a two-layer FC neural network with $m$ hidden neurons:
\begin{align*}
   f(\x;\W,\z) &= \frac{1}{\sqrt{m}}\sum_{r=1}^m z_r \phi(\w_r^T\x/\sqrt{d}) \\
   &= \frac{1}{\sqrt{m}}\z^T\phi(\W\x/\sqrt{d}),
\end{align*}
where $\x \in \R^d$ is the input, $\W = [\w_1, \cdots , \w_m]^T \in \R^{m\times d}$
is the weight matrix in the first layer, and
$\z = [\z_1, \cdots , \z_m]^T\in \R^m$ is the weight vector in the second layer. Here $\phi: \R \rightarrow \R$ is a smooth or piece-wise linear activation function (including ReLU, Leaky ReLU, Erf,
Tanh, Sigmoid, Softplus, etc.) that acts entry-wise on vectors or matrices.
We consider the following $\ell_2$ training loss: \vspace{-1mm}
\begin{equation}\label{eq:l2}
    \LL(\W,\z)=\frac{1}{2n}\sum_{i=1}^n (f(\x_i;\W,\z)-y_i)^2.
\end{equation}
We train the network by applying gradient descent on the loss \eqref{eq:l2} starting from random initialization\footnote{Detailed assumptions on the %
activations, and initialization can be found in \cref{apdx:setting}}\looseness=-1:
\begin{align}
    \W_{t + 1} &= \W_t - \eta \nabla_{\W} \LL(\W_t, \z_t), \\  
    \z_{t + 1} &= \z_t - \eta \nabla_{\z} \LL(\W_t, \z_t), 
\end{align}

\paragraph{Worst-group Error.} We quantify the performance of the model based on its highest test error across groups $\mathcal{G}=\{g_{c,s}\}_{c,s}$ in all classes. Formally, \textit{worst-group} test error is defined as:
\begin{equation}
    \text{Err}_{wg}=\max_{g\in \mathcal{G}}\E_{(\x_i,y_i)\in g}[
    y_i\neq y_f(\x_i;\W,\z)
    ], 
\end{equation}
where $y_f(\x_i;\W,\z)$ is the label predicted by the model.
That is, $\text{Err}_{wg}$ measures the highest fraction of examples that are incorrectly classified across all groups.
\looseness=-1

\section{INVESTIGATING SPURIOUS
FEATURE LEARNING IN
NEURAL NETWORKS}\label{sec:theory} 
We start by investigating how spurious features are learned during training a two-layer fully-connected neural network. 
Our analysis reveals two phases in early-time learning. First, in the initial training iterations, the contribution of a spurious feature %
to the network output increases linearly with the amount of the spurious correlation. %
Interestingly, if the majority group is sufficiently large, majority and minority groups are separable at this phase by the network output.
Second, if the noise-to-signal ratio of the spurious feature of the majority group is smaller than that of the core feature, the network's output on the majority of examples in the class will be almost exclusively determined by the spurious feature and will remain mostly invariant to the core feature. 
Next, we will discuss the two phases in detail.\looseness=-1

\begin{figure}[t]
     \centering
\includegraphics[width=\columnwidth]{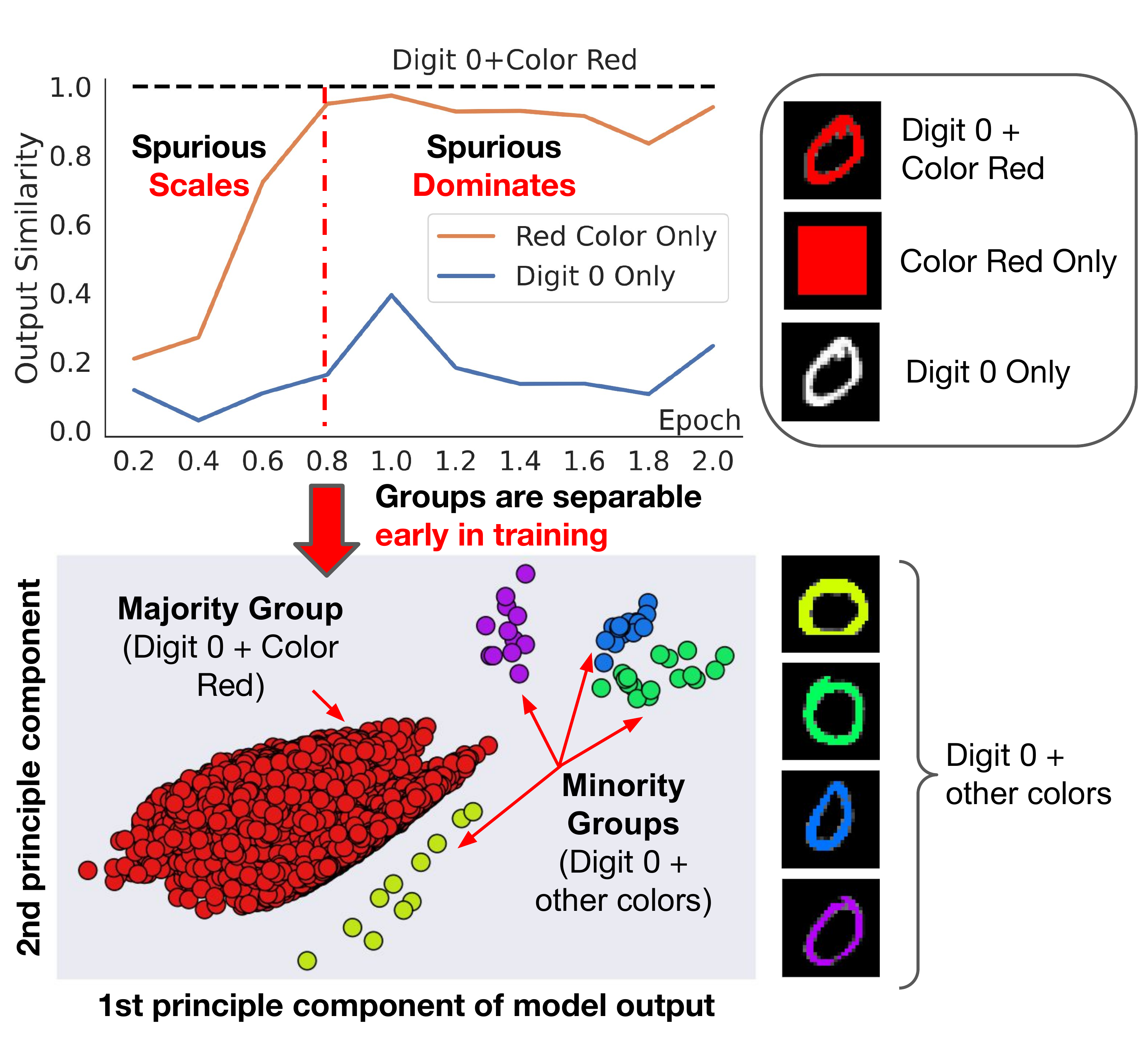}
\vspace{-2mm}
     \caption{Training LeNet-5 on Colored MNIST. \textbf{Top}: Up to epoch 2, the network output is almost exclusively indicated by the color red (spurious feature in the majority group).
     \textbf{Bottom}: Majority and minority groups are separable based on the network output, e.g. via clustering. Minority groups that have a spurious feature in majority groups of other classes (yellow, purple, blue, green) are also separable from each other.
     Similar results on Waterbirds are shown in \cref{fig:rebuttal-pred-diff}. 
     \looseness=-1}\label{fig:pred-diff}
\end{figure}

\subsection{Spurious Features are Learned in the Initial Training Iterations}\label{sec:mi}
We start by analyzing the effect of spurious features on learning dynamics of a two-layer FC neural network trained with gradient descent in the initial training iterations. 
{With the data model $\x_i=\vct{v}_{c}+\vct{v}_{s}+\vct{\xi}_i$ defined in \cref{sec:problem},} the following theorem shows that if a majority group %
is sufficiently large, %
contribution of its spurious feature %
to the model's output is magnified by the network %
at every step early in training.
\begin{restatable}{theorem}{linparam}\label{thm:lin_param}
Let $\alpha \in (0,\frac{1}{4})$ be a fixed constant. Suppose the number of training samples $n$ and the network width $m$ satisfy $n \gtrsim d^{1+\alpha}$ and $m \gtrsim d^{1+\alpha}$.
Let $n_{c}$ be the number of examples in class $c$, and $n_{c,s}\!=\!|g_{c,s}|$ be the size of group $g_{c,s}$ with label $c$ and spurious feature $\vct{v}_{s}\in\A$.
Then, under the setting of \cref{sec:problem} there exist a constant $\nu_1>0$, such that with high probability, for all $0\leq t \leq \nu_1\cdot \sqrt{\frac{d^{1-\alpha}}{\eta}}$,
the contribution of the core and spurious features to the network output can be quantified as follows:\looseness=-1 
\begin{align} 
    f(\vb_c;\W_t,\z_t) &= 
    \frac{2\eta \zeta^2 c \|\vct{v}_c\|^2 t}{d}  \left(\frac{n_c}{n} \pm \OO(d^{-\Omega(\alpha)})\right), \\
   f(\vb_s;\W_t,\z_t) \!&=\! \frac{2 \eta \zeta^2 c \|\vct{v}_s\|^2 t}{d} \bigg(\frac{n_{c,s} - n_{c',s}}{n} \\
   \nonumber
   & \qquad\qquad\qquad\qquad \pm \OO(d^{-\Omega(\alpha)})\bigg) , 
\end{align}
where %
$c'\!=\mathcal{C}\!\setminus\! c$, and $\zeta$ is the expected gradient of activation functions at random initialization. 
\end{restatable}
The proof for \cref{thm:lin_param} is detailed in \cref{apdx:proof_early_dynamics}. 
At a high level, as the model is nearly linear in the initial $\nu_1\cdot\frac{d \log d}{\eta}$ iterations, the contribution of the spurious feature $\vb_s$ to the network output grows almost {linearly} with $(n_{c,s} \!- n_{c',s})\|\vb_s\|^2$, at \textit{every iteration} in the initial phase of training. 
Here, $n_{c,s} \!- n_{c',s}$ represents the correlation between the spurious feature and label $c$.
When $n_{c,s}\gg n_{c',s}$, the spurious feature exists almost exclusively in the majority group of class $c$, and thus has a high correlation only with class $c$. In this case, if the magnitude of the spurious feature is significant, the contribution of the spurious feature to the model's output grows very rapidly, early in training. 
In particular, if $(n_{c,s} \!- n_{c',s})\|\vb_s\|^2\gg n_c\|\vb_c\|^2$, the model's output is increasingly determined by the spurious feature, but not the core feature.

Remember from \cref{sec:problem} that every example consists of a core and a spurious feature. 
As the effect of spurious features of the majority groups is amplified in the network output, the model's output will differ for examples in the majority and minority groups.
The following corollary shows that the majority and minority groups are separable based on the network's output early in training.
Notably, multiple minority groups with spurious features contained in majority groups of other classes are also separable. \looseness=-1

\begin{restatable}[\textbf{Separability of majority and minority groups}]{corollary}{cluster}\label{thm:cluster}
Suppose that for all classes, a majority group has at least $K$ examples and a minority group has at most $k$ examples. Then, under the assumptions of \cref{thm:lin_param}, examples in the majority and minority groups are {nearly} separable {with high probability} based on the model's output, early in training. That is, for all $0\leq t \leq \nu_1\cdot \sqrt{\frac{d^{1-\alpha}}{\eta}}$, with high probability, the following holds for at least $1 - \OO(d^{-\Omega(\alpha)})$ fraction of the training examples
$\vct{x}_i$ in group $g_{c,s}$:
    
If $g_{c,s}$ is in a majority group in class $c=1$: %
    \begin{align}
        f(\x_i;\W_t,\z_t) &\geq \frac{2\eta \zeta^2 t}{d} \bigg(\frac{\|\vct{v}_{s}\|^2 (K - k)}{n}  \\
        \nonumber
        &+ \xi \pm \OO(d^{-\Omega(\alpha)})%
        \bigg) + \rho(t,\phi,\Sigma),
    \end{align}
If $g_{c,s}$ is in a minority group in class $c=1$, but $g_{c',s}$ is a majority group in class $c'=-1$:
    \begin{align}
        f(\x_i;\W_t,\z_t) &\leq \frac{2\eta \zeta^2 t}{d} \bigg(-\frac{\|\vct{v}_{s}\|^2 (K - k)}{n}  \\ 
        \nonumber
        &+ \xi \pm \OO(d^{-\Omega(\alpha)})%
        \bigg) + \rho(t,\phi,\Sigma),
    \end{align}

where $\rho$ is constant for all examples in the same class,
$\xi\sim\mathcal{N}(0, \kappa)$ with $\kappa = \frac{1}{n}(\sum_{c} n_{c}^2 \sigma_c^2 \|\vct{v}_c\|^2)^{1/2} + \frac{1}{n}(\sum_{s} (n_{c,s} \!-\! n_{c',s})^2 \sigma_s^2 \|\vct{v}_s\|^2)^{1/2}$ 
is the total effect of noise on the model. %

Analogous statements holds for the class $c\!=\!-1$ by changing the sign and direction of the inequality.\looseness=-1
\end{restatable}

The proof can be found in \cref{apdx:proof_early_dynamics}. 
\cref{thm:cluster} shows that when the majority group %
is considerably larger than the minority groups ($K \gg k)$, the prediction of examples in the majority group 
move toward their label considerably faster, due to the contribution of the spurious feature.
Hence, majority and minority groups can be separated from each other, early in training. Importantly, multiple minority groups can be also separated from each other, if their spurious feature exists in majority groups of other classes.
Note that $K>k+|\xi|$ is the minimum requirement for the separation to happen. Separation is more significant when $K\gg k$ and when $\|\vb_s\|$ is significant. %

\subsection{Network Relies on Simple Spurious Features for Majority of Examples}
Next, we analyze the second phase in early-time learning of a two-layer neural network. In particular, we show that if the noise-to-signal ratio of the spurious feature of the majority group of class $c$, i.e., $R_s={\sigma_s}/{\|\vb_s\|}$ is smaller than that of the core feature $R_c={\sigma_c}/{\|\vb_c\|}$, 
then the neural network's output is almost exclusively determined by the spurious feature and remain invariant to the core feature at $T\!=\nu_2\cdot\frac{d \log{d}}{\eta}$, even though the core feature is more predictive of the class. 

\begin{restatable}{theorem}{optimallinparam} \label{thm:optimal_lin_param}
Under the assumptions of \cref{thm:lin_param}, %
if the classes are balanced, and the total size of the minority groups in class $c$ is small, i.e., $\OO(n^{1-\gamma})$ for some $\gamma>0$,
then %
there exists a constant $\nu_2 > 0$ such that at $T=\nu_2\cdot\frac{d\log d}{\eta}$, for an example $\x_i$ in a majority group $g_{c,s}$, the contribution of the core feature to the model's output is at most: %
\begin{align}
    |f(\vb_c;\W_{T},\z_T)| &\leq 
    \left(\frac{\sqrt{2} R_s}{R_c} + \OO(n^{-\gamma} + d^{-\Omega(\alpha)}) \right).
\end{align}
In particular if %
$\min\{R_c, 1\} \gg R_s$, then the model's output is mostly indicated by the spurious feature instead of the core feature: \vspace{-1mm}
\begin{align}
    |f(\vb_s;\W_T,\z_{T})| \gg |f(\vb_c;\W_T,\z_{T})|. \vspace{-2mm}
\end{align} 
\end{restatable}\vspace{-2mm}
The proof can be found in \cref{apdx:proof_optimal_lin_param}. 
\cref{thm:optimal_lin_param} 
shows that at $T=\nu_2\cdot \frac{d \log d}{\eta}$, 
the contribution of the core feature to the network's output is at most proportional to $R_s/R_c$. Hence, if $R_s\ll R_c$, 
the network almost exclusively relies on the spurious feature of the majority group instead of the core feature.\looseness=-1

We note that our results in \cref{thm:lin_param}, \cref{thm:cluster}, and \cref{thm:optimal_lin_param} generalize to more than two classes 
and hold if the classes are imbalanced, as we will confirm by our experiments. Similar results can be shown for multi-layer fully connected and convolutional networks, following \citep{hu2020surprising}. %

\subsection{Separability of Majority and Minority Groups in The Two Early Training Phases}
The initial phase of training iterations, 
when $0\leq t \leq \nu_1\cdot \sqrt{\frac{d^{1-\alpha}}{\eta}}$, is characterized by
approximately linear change in the loss. 
\cref{thm:cluster} shows that in Phase 1,
majority and minority groups are separable based on the network output,
if spurious feature is strongly correlated with label and %
has a higher magnitude than the core.
Phase 2 happens when $T=\nu_2\cdot \frac{d \log d}{\eta}$
and marks the point where the approximate linear model converges to its optimal parameters, and the network starts learning higher-order (non-linear) functions. 
\cref{thm:optimal_lin_param} shows that in Phase 2, %
majority and minority groups are separable based on the network output,
if the noise-to-signal ratio of spurious feature ($R_s$) is smaller than core ($R_c$). %

The above discussion implies that %
one can separate majority and minority groups in Phase 1 or Phase 2, as long as the corresponding conditions are met. %

\vspace{-1mm}
\paragraph{Visualization of Theoretical Results.} We empirically illustrate the above results 
during early-time training of LeNet-5 \citep{lecun1998gradient} on the Colored MNIST \citep{alain2015variance} dataset containing colored handwritten digits. 
\cref{fig:pred-diff} shows that the prediction of the network on the majority group is almost exclusively indicated by the color of the majority group, confirming \cref{thm:optimal_lin_param}. 
The bottom of \cref{fig:pred-diff} shows that the majority and minority groups are separable based on the network output, confirming \cref{thm:cluster}.\looseness=-1

\vspace{-1mm}
\paragraph{Strong Spurious Correlations Make the Network Invariant to Core Features of Majority Groups.}
Finally, note that by only learning the spurious feature, the neural network can %
shrink the training loss on the majority of examples in class $c$ to nearly zero and correctly classify them. 
Hence, the contribution of the spurious feature of the majority group of class $c$ to the model's output is retained throughout the training. On the other hand, %
if minority groups are small, %
higher complexity functions that appear later in training overfit the minority groups, as observed by \citep{sagawa2020investigation}.
This results in a small training error but a poor worst-group generalization performance on the minorities.\looseness=-1 %

\begin{algorithm}[t]
  \caption{\fullname (\alg)}\label{alg:alg}
  \begin{algorithmic}
    \REQUIRE {\!Network $f(.,\W)$, data $\mathcal{D}=\{(\x_i,y_i)\}_{i=1}^n$, 
    loss function $\LL$, 
    iteration numbers $T_N,T_{init}$.
    }\looseness=-1
    \ENSURE{Model $f$ trained without bias}
    \STATE \textbf{Stage 1: Early Bias Identification}
    \FOR {$t=0, \cdots, T_{init}$}  
    \STATE $\W_{t+1}\leftarrow \W_t-\eta\nabla\LL(\W_t;\D)$
    \ENDFOR
    \FOR {every class $c\in\mathcal{C}$ with examples $V_c$}
    \STATE Identify $\lambda$, \# of clusters $k$ via Silhouette analysis %
    \STATE Cluster $V_c$ into $\{V_{c,j}\}_{j=1}^k$ based on $\!f(\x_i;\W_t)$
    \STATE Weight every $\x_i\!\!\in\!\! V_{c,j}$ by $w_i \!\!=\!\! 1/|V_{c,j}|$, $p_i = w_i^{\lambda}/{\sum_{i} w_i^{\lambda}}$ 
    \ENDFOR
    \STATE \textbf{Stage 2: Learning without Bias}
    \FOR{$t = 0,\cdots, T_N$}
    \STATE Sample a mini-batch $\mathcal{M}_t\!=\!\{(\x_i,y_i)\}_i$ with probabilities ${p_i}$
    \STATE $\W_{t+1}=\W_t-\eta\nabla\LL(\W_t;\mathcal{M}_t)$.
    \ENDFOR
  \end{algorithmic}
\end{algorithm}

\section{SPARE: ELIMINATING
SPURIOUS BIAS EARLY IN
TRAINING}
\label{sec:method}
Based on the theoretical foundations outlined in \cref{sec:theory}, we develop a principled pipeline, \alg, to discover and mitigate spurious correlations \textit{early in training}. The pseudocode %
is illustrated in \cref{alg:alg}.\looseness=-1

\vspace{-1mm}
\paragraph{Discovering Spurious Correlations: Separating the Groups Early in Training.} 
\cref{thm:cluster} shows that majority and minority groups are separable based on the network's output. To identify the majority and minority groups, we cluster examples $V_c$ in every class $c\in\mathcal{C}$ %
based on the output of the network, during the first few epochs. We tune $T_{init}$ for maximum recall of \alg's clusters against the validation set groups in the first 1-2 epochs (discussed in \cref{sec:hyper}). 
We determine the number of clusters via silhouette analysis \citep{rousseeuw1987silhouettes}, a technique that assesses the cohesion and separation of clusters by evaluating how close each point in one cluster is to points in the neighboring clusters. 
In doing so, we can separate majority and minority groups in each class of examples with different spurious features.
Any clustering algorithm such as $k$-means or $k$-median clustering \citep{mirzasoleiman2013distributed,mirzasoleiman2015lazier} can be applied to separate groups in large data.\looseness=-1

\vspace{-1mm}
\paragraph{Mitigation after Discovery: Balancing Groups via Importance Sampling.}
To alleviate the spurious correlations and enable effective learning of the core features, we employ an importance sampling method on examples in each class to upsample examples in the smaller clusters and downsample examples in the larger clusters. 
To do so, we assign every example $i\in V_{c,j}$ a weight given by the size of the cluster it belongs to, i.e., $w_i=1/|V_{c,j}|$. %
Then we sample examples in every mini-batch with probabilities equal to $p_i = w_i^{\lambda}/{\sum_{i} w_i^{\lambda}}$, where %
$\lambda$ can be determined based on the average silhouette score of clusters in each class, \textit{without further tuning}. 
Our importance sampling method does not increase the size of the training data but only changes the data distribution. Hence, it does not increase the training time.\looseness=-1

%% file: sections/4_experiments.tex
\begin{table*}[t]
\caption{Worst-group  and average accuracy ($\%$) of training with \alg vs. state-of-the-art algorithms. %
\alg achieves a superior performance much faster that existing methods. 
CB, GB indicate balancing classes and groups. 
Range for training cost encompasses all datasets, and accounts for (1) training the reference model for group inference and (2) 
number of training examples involved in robust training (excluding tuning cost). 
Baseline results for CMNIST, UrbanCars are %
from the benchmarks %
\citep{zhang2022correct,li2023whac}.
$\blacklozenge$ and $\triangle$ indicate using group-labeled validation for tuning group inference, and robust training. (E\#) shows the early group inference epoch for \alg. $(\lozenge)$ shows \alg doesn't heavily rely on validation set $(\lozenge)$. We couldn't successfully run CnC on UrbanCars and DFR on CMNIST. \looseness=-1
}  
\vspace{-2mm}
\label{tab:spurious}
\begin{center}
\resizebox{\textwidth}{!}{%
\begin{tabular}{lrrrrrrrrrrrrrr}
\toprule
 & Grp label & Train & \multicolumn{2}{c}{\bf CMNIST}  &\multicolumn{2}{c}{\bf Waterbirds}   &\multicolumn{2}{c}{\bf CelebA} &\multicolumn{2}{c}{\bf UrbanCars} \\
  & required & cost & \multicolumn{2}{c}{(1 Spurious $\times$ {\bf 5 Classes})}  &\multicolumn{2}{c}{(1 Spurious $\times$ 2 Classes)}   &\multicolumn{2}{c}{(1 Spurious $\times$ 2 Classes)}  &\multicolumn{2}{c}{({\bf 2 Spurious} $\times$ 2 Classes)}  \\
 \cmidrule(lr){4-5}
 \cmidrule(lr){6-7}
 \cmidrule(lr){8-9}
 \cmidrule(lr){10-11}
 & & & Worst-group & Average & Worst-group & Average & Worst-group & Average & Worst-group & Average \\
 \midrule\midrule
 ERM & $-$$-$ & 1x & $0.0_{\pm 0.0}$ & $20.1_{\pm 0.2}$ & $62.6_{\pm 0.3}$ & $97.3_{\pm 1.0}$ & $47.7_{\pm 2.1}$ & $94.9_{\pm 0.3}$ & $28.4$ & $97.6$ \\ 
 CB & $-$$-$ &  1x & $0.0_{\pm 0.0}$ & $23.7_{\pm 3.1}$  & $62.8_{\pm 1.6}$ & $97.1_{\pm 0.1}$ & $46.1_{\pm 1.5}$ & $95.2_{\pm 0.4}$ & 33.7 & 98.1 \\ 
 \textsc{George} \citep{sohoni2020no} & $-$$-$ & 2x & $76.4_{\pm 2.3}$ & $89.5_{\pm 0.3}$ & $76.2_{\pm 2.0}$ & $95.7_{\pm 0.5}$ & $54.9_{\pm 1.9}$ & $94.6_{\pm 0.2}$  & 35.2 & 97.9 \\
 PGI \citep{ahmed2020systematic} & $\blacklozenge$ - & 1x & $73.5_{\pm 1.8}$ & $88.5_{\pm 1.4}$ & $79.5_{\pm1.9}$ & $95.5_{\pm 0.8}$ & $85.3_{\pm 0.3}$ & $87.3_{\pm 0.1}$  & 34.0 & 95.7 \\ 
 EIIL \citep{creager2021environment} & $\blacklozenge$ - & 1x & $72.8_{\pm {\bf 6.8}}$ & $90.7_{\pm 0.9}$  & $83.5_{\pm 2.8}$ & $94.2_{\pm 1.3}$ & $81.7_{\pm 0.8}$ & $85.7_{\pm 0.1}$ & $50.6$ & $95.5$\\
 LfF \citep{nam2020learning}&  $\blacklozenge\triangle$ &  2x & $0.0_{\pm 0.0}$ & $25.0_{\pm 0.5}$ & $78.0_{N/A}$ &  $91.2_{N/A}$ & $77.2_{N/A}$ & $85.1_{N/A}$ & 34.0 & 97.2 \\
 JTT \citep{liu2021just} & $\blacklozenge\triangle$ & 5x-6x &  $74.5_{\pm {\bf 2.4}}$ & $90.2_{\pm 0.8}$ & $83.1_{\pm {\bf 3.5}}$ &  $90.6_{\pm 0.3}$ & $81.5_{\pm 1.7}$ & $88.1_{\pm 0.3}$ & 55.8 & 95.9 \\
 CnC \citep{zhang2022correct} & $\blacklozenge\triangle$ & 2x-12x & $77.4_{\pm 3.0}$ & $90.9_{\pm 0.6}$ & $88.5_{\pm 0.3}$ &  $90.9_{\pm 0.1}$ & $88.8_{\pm 0.9}$ & $89.9_{\pm 0.5}$ & - & -\\
 \rowcolor{lightgray!40} \textbf{\alg} & $\lozenge$ - & \textbf{1x} & \textbf{(E2)} $\mathbf{83.0_{\pm 1.7}}$ & $91.8_{\pm 0.7}$ & \textbf{(E2)} $\mathbf{91.6_{\pm 0.8}}$ & $96.2_{\pm 0.6}$ & \textbf{(E1)} $\mathbf{90.3_{\pm 0.3}}$ & $91.1_{\pm 0.1}$ & \textbf{(E2)} $\mathbf{76.9_{\pm 1.8}}$ & $96.6_{\pm 0.5}$ \\
  \rowcolor{lightgray!40} \multicolumn{2}{l}{($-2^{nd}$ best)}   &  & ($+ \mathbf{5.6}$) &  &  ($+ \mathbf{3.1}$) &  &  ($+ \mathbf{1.5}$) &  &  ($+ \mathbf{21.1}$) &  \\
 \midrule\midrule
 $\text{DFR}$ \citep{kirichenko2023last} & train sub & 1x & - & - & $90.4_{\pm 1.5}$ & $94.1_{\pm 0.5}$ & $80.1_{\pm 1.1}$ & $89.7_{\pm 0.4}$ & 44.5 & 89.7 \\
 GB & train full & 1x & $82.2_{\pm 1.0}$ & $91.7_{\pm 0.6}$ & $86.3_{\pm 0.3}$ & $93.0_{\pm 1.5}$ & $85.0_{\pm 1.1}$ & $92.7_{\pm 0.1}$ & 73.9 & 92.2 \\
 GDRO \citep{sagawa2019distributionally} &  train full & 1x & $78.5_{\pm 4.5}$ & $90.6_{\pm 0.1}$  &  $89.9_{\pm 0.6}$ &  $92.0_{\pm 0.6}$ & $88.9_{\pm 1.3}$ & $93.9_{\pm 0.1}$ & 75.2 & 91.6 \\
\bottomrule
\end{tabular}%
}
\end{center}
\end{table*}

\section{EXPERIMENTS}
In this section, we first confirm that \alg outperforms state-of-the-art baselines in identifying and mitigating spurious correlations across multiple curated benchmark datasets. Most notably, \alg excels on UrbanCars, a challenging dataset with multiple spurious correlations within each class. Then, we demonstrate that \alg effectively discovers and mitigates naturally occurring spurious correlations early in training on Restricted ImageNet—a realistic dataset \textit{without known spurious correlations}. \looseness=-1

\subsection{Mitigating Curated Spurious Correlations in Benchmark Datasets}
First, we evaluate the effectiveness of \alg in alleviating spurious correlations on spurious benchmarks. The reported results are averaged over three runs with different model initializations.\looseness=-1

\vspace{-1mm}
\paragraph{Benchmark Datasets \& Models.}
(1) {CMNIST} \citep{alain2015variance} contains colored handwritten digits derived from MNIST \citep{lecun1998gradient}. 
We follow the challenging 5-class setting in \cite{zhang2022correct} where every two digits form one class and 99.5\% of training examples in  each class are spuriously correlated with a distinct color. We use a 5-layer CNN (LeNet-5 \citep{lecun1998gradient}) for CMNIST. 
(2) {Waterbirds} %
\citep{sagawa2019distributionally} contains two classes (landbird vs. waterbird) and the background (land or water) is the spurious feature. Majority groups are (waterbird, water) and (landbird, land).
(3) {CelebA} \citep{liu2015faceattributes} is a face datasets containing two classes (blond vs dark) %
and gender (male or female) as the spurious feature \citep{sagawa2019distributionally}. Majority groups are (blond, female) and (non-blond, male).
(4) UrbanCars \citep{li2023whac} %
is a challenging task
containing two classes
(urban vs. country cars) and \textit{two spurious features}: (1) background (BG): urban vs. country and (2) co-occurring object (CoObj): fireplug and stop sign vs cow and horse. 
For Waterbirds, CelebA, and UrbanCars, we follow the standard settings in previous work to train a ResNet-50 model \citep{he2016deep} pretrained on ImageNet provided by the Pytorch library \citep{pytorch}. \looseness=-1
More details about the datasets and the experimental settings are in \cref{sec:experiment}.

\vspace{-2mm}
\paragraph{Baselines.}
We compare \alg with the state-of-the-art methods for eliminating spurious correlations in \cref{tab:spurious}, in terms of both {worst-group} %
and average accuracy. We use {adjusted average accuracy} for Waterbirds, i.e., the average accuracy over groups weighted by their size. This is consistent with prior work, and is done because the validation and test sets are group-balanced while the training set is skewed. For UrbanCars, our average accuracy corresponds to the ID accuracy in the original paper \citep{li2023whac}, and our worst-group accuracy is computed by adding the largest of BG/CoObj/BG+CoObj gap to the ID accuracy.
GB (Group Balancing) and GDRO \citep{sagawa2019distributionally} use the group label of all training examples.
DFR \citep{kirichenko2023last} %
uses group-balanced data
drawn from 
either validation ($\text{DFR}^{Tr}_{Tr}$) or 
training 
($\text{DFR}^{Val}_{Tr}$) 
data. {We considered DFR$_{tr}^{tr}$, which trains on group-balanced training data, for a fair comparison with baselines that only use the validation set to tune.
The rest of the methods infer the group labels without using such information.\looseness=-1

\begin{table*}[ht]
\begin{minipage}[b]{0.4\textwidth}
\centering
\resizebox{\textwidth}{!}{%
\begin{tabular}{lrrr}
    \toprule
    \multicolumn{1}{c}{\bf Method} & \multicolumn{1}{c}{\bf BG ($\uparrow$)}	& \multicolumn{1}{c}{\bf CoObj ($\uparrow$)}	& \multicolumn{1}{c}{\bf BG+CoObj ($\uparrow$)} \\
    \midrule
    JTT (E1) & -8.1 & \textcolor{red}{-13.3} & \textcolor{red}{-40.1} \\
    EIIL (E1) & -4.2 & \textcolor{red}{-24.7} & \textcolor{red}{-44.9} \\
    JTT (E2) & \textcolor{red}{-23.3} & -5.3 & \textcolor{red}{-52.1} \\
    EIIL (E2) & \textcolor{red}{-21.5} & -6.8 & \textcolor{red}{-49.6} \\
    \rowcolor{lightgray!40} \textbf{\alg} (E2) & \textbf{-5.3} & \textbf{-3.1} & \textbf{-8.9} \\
    \bottomrule
   \end{tabular}%
    }
    \caption{UrbanCars with two spurious correlations (BG and CoObj). SOTA methods show ``whack-a-mole'' behavior \citep{li2023whac}: mitigating one spurious correlation amplifies the other, 
    \alg finds minority groups more accurately and %
    does not exhibit whack-a-mole.
    \looseness=-1}
    \label{tab:whac-a-mole}
\end{minipage}\hfill
\begin{minipage}[b]{0.57\textwidth}
     \centering
    \begin{subfigure}[b]{0.24\textwidth}
         \centering
         \includegraphics[width=\textwidth]{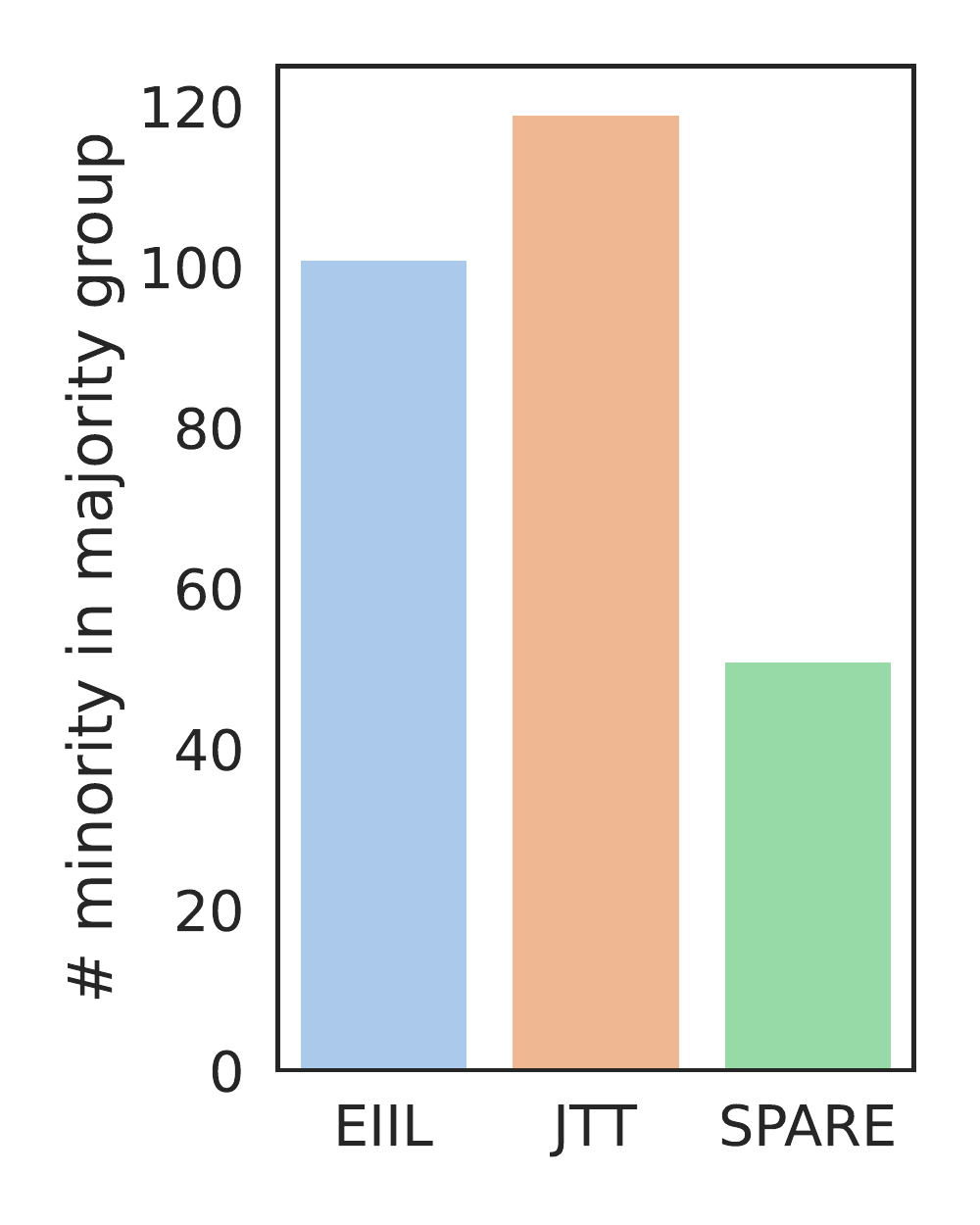}
         \vspace{-2em}
         \caption{Waterbirds}
         \label{fig:waterbirds-minor-in-major}
     \end{subfigure}
     \begin{subfigure}[b]{0.24\textwidth}
         \centering
        \includegraphics[width=\textwidth]{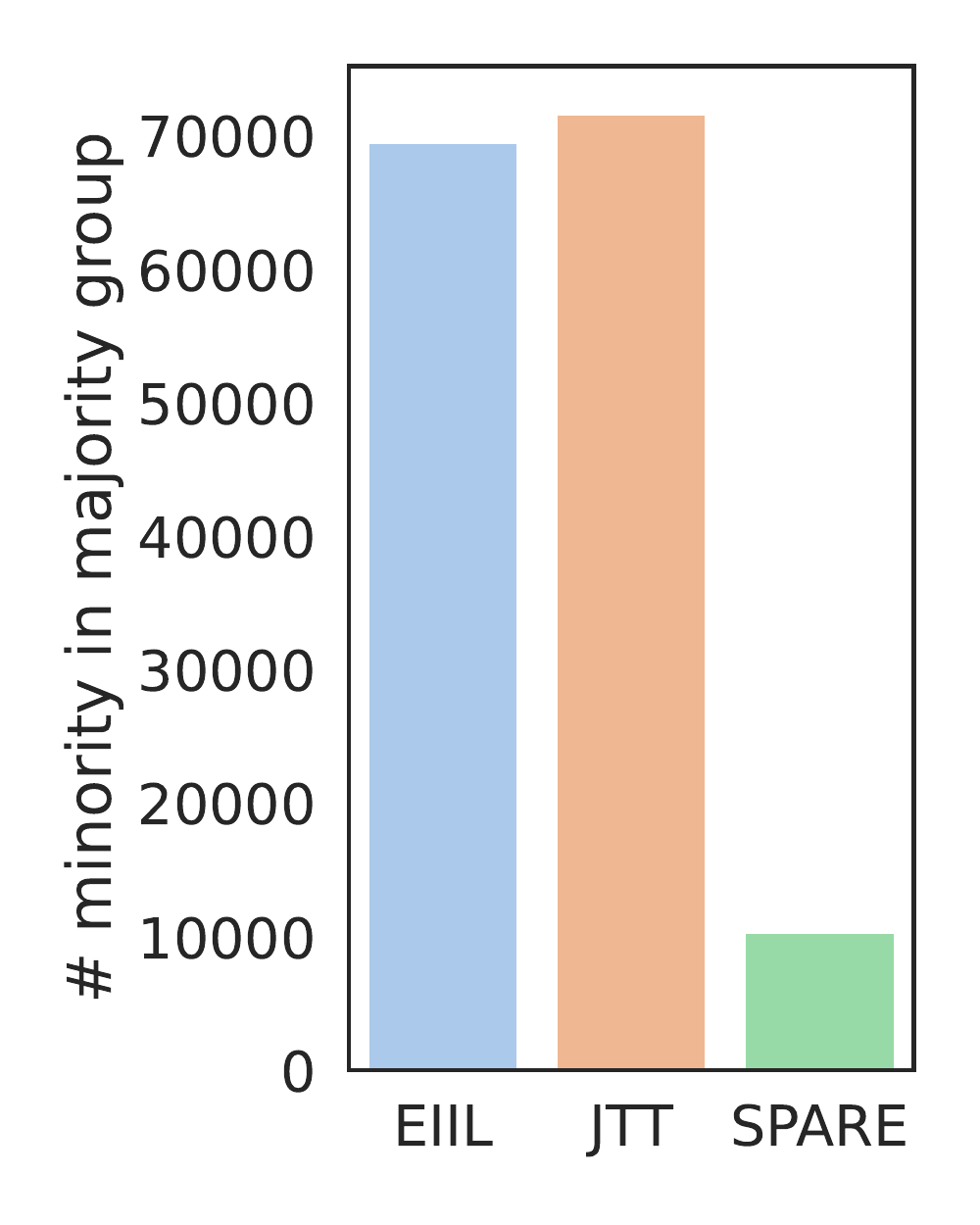}
        \vspace{-2em}
         \caption{CelebA}
         \label{fig:celeba-minor-in-major}
     \end{subfigure}
     \begin{subfigure}[b]{0.48\textwidth}
         \centering
         \includegraphics[width=\textwidth]{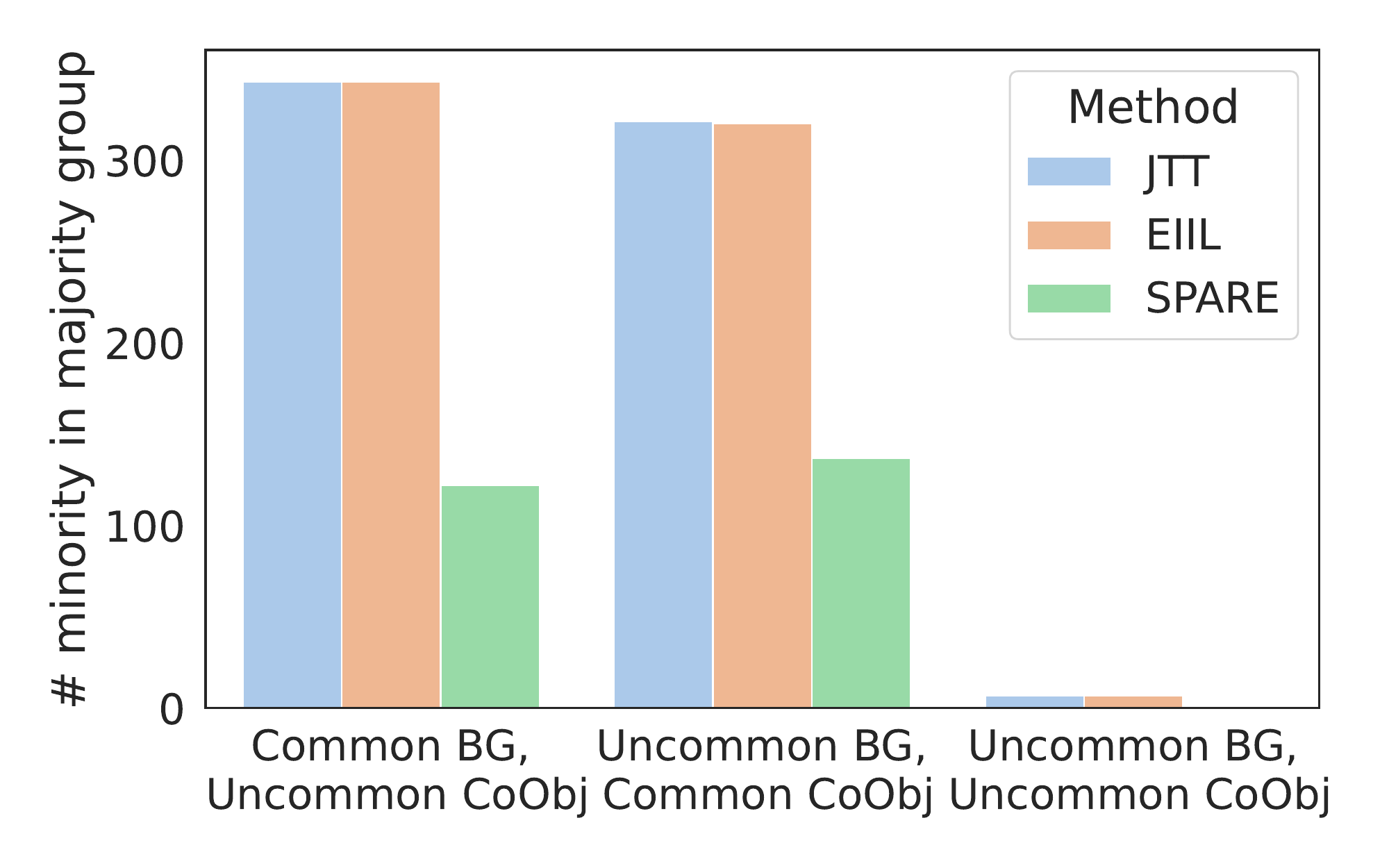}
         \vspace{-2em}
         \caption{UrbanCars}
         \label{fig:urbancars-minor-in-major}
     \end{subfigure}
     \captionof{figure}{Number of minority examples inferred as majority. JTT and EIIL infer many minority examples as majority and mistakenly downweight them. \alg identifies minority groups more accurately, and correctly upweights them.} \looseness=-1\label{fig:minor-in-major}
    \vspace{-1mm}
\end{minipage}
\end{table*}

\vspace{-2mm}
\paragraph{\alg Outperforms SOTA Algorithms, Especially When Multiple Spurious Correlations Exist.} \cref{tab:spurious} shows that 
\alg consistently outperforms the best baselines, 
on worst-group and average accuracy. 
On UrbanCars, where baselines have been shown to amplify one spurious when trying to mitigate the other \citep{li2023whac}, \alg achieves a 21.1\% higher accuracy than the next best state-of-the-art method that does not rely on ground-truth group labels during training.
Notably, \alg performs comparably to those that \textit{use} the group information, and even achieves a better worst-group accuracy on CMNIST, CelebA, and UrbanCars, and has a comparable worst-group but higher average accuracy on the Waterbirds. 
As group labels are unavailable in real-world datasets, methods that do not rely on group labels are more practical. Moreover, group labels can sometimes be inaccurate, so group-inference methods like \alg can better identify the groups.
This aligns with EIIL's observation that inferred groups can be more informative than underlying group labels \citep{creager2021environment}.
Among such methods, \alg has a superior performance and easily scales to large datasets.

\vspace{2mm}
\paragraph{\alg Does Not Exhibit Whack-a-mole Behavior when Data Contains Multiple Spurious Correlations.} \cref{tab:whac-a-mole} shows the accuracy of different groups in UrbanCars datasets with two spurious correlations (BG and CoObj). We see that JTT and EIIL drop the accuracy on BG or CoObj when trying to improve the accuracy on the other. This is because they find a smaller minority group, and hence need to upweight it heavily to mitigate the spurious correlation. In doing so, they amplify the other subtle spurious correlation in the minority group. In contrast, \alg finds minority groups more accurately and mitigates the spurious correlation without introducing a new one. Notably, \alg achieves 31.2\% better accuracy than the best baseline on the smallest group (BG+CoObj).

\vspace{-2mm}
\paragraph{\alg is Much Faster and Easier to Tune.} As our theoretical results narrow the range for inference time to the initial epochs (often 1 or 2 in practice), it can be tuned easily for \alg, while others need to search over a wide range from epoch 1 to 60. In addition to the time saved for hyperparameter tuning, \alg 
has
up to 12x lower computational cost ($k$-means \& total wall-clock runtimes are reported in \cref{tab:cluster} and \cref{tab:wallclock}) compared to the state-of-the-art.\looseness=-1

\subsection{Why Does \alg Work Better?}\label{sec:better}
\paragraph{\alg Finds Minority Groups More Accurately.} \cref{fig:minor-in-major} explains the superior performance of \alg over the state-of-the-art: JTT \citep{liu2021just} and EIIL \citep{creager2021environment} mistakenly group many minority examples into majority groups %
and thus mistakenly downweight %
them. {The minority samples include, by definition, all instances where spurious correlations are not present.}
In contrast, by finding the minority groups more accurately, \alg effectively upweights them to mitigate the spurious correlations and improve the worst-group accuracy.
We also compare the worst-group and average accuracy of models trained with GDRO and JTT's upsampling method applied to groups inferred by \alg in \cref{tab:sampling}. Comparing to \cref{tab:spurious}, we see that both methods obtain a better worst-group accuracy using \alg's groups. In particular, training on \alg's groups with GDRO outperform George and EIIL that use GDRO by 11.4\%, 4.1\%. Similarly, training on \alg's groups with JTT's upsampling outperforms JTT by 4.5\%, further confirming that \alg finds minorities more accurately.

\vspace{-1mm}
\paragraph{\alg's Importance Sampling is More Effective and Efficient.} Next, we compare the worst-group and average accuracy of models trained with GDRO, JTT's upsampling, and \alg's importance sampling, applied to groups inferred by \alg. \cref{tab:sampling} shows that  \alg's importance sampling is more effective in improving the worst-group accuracy and outperforms GDRO by 4\% and JTT's upsampling by 5.4\%. Note that both methods require tuning based on group-labeled validation data, and upsampling drastically increases the training time. On the other hand, \alg's importance sampling does not require any hyperparameter tuning or increase the training time.\looseness=-1

\begin{table}
    \label{tab:ablation}
    \caption{\alg's importance sampling is more effective in improving the worst-group accuracy than GDRO and JTT's upsampling, when applied to \alg's groups. 
    }
        \vspace{-3mm}
    \label{tab:sampling}
    \centering
    \resizebox{\columnwidth}{!}{%

    \label{tab:sample}
    \centering
    \begin{tabular}{cccc}
    \toprule
 \multicolumn{1}{c}{\bf Groups} & \multicolumn{1}{c}{\bf Robust training} & \multicolumn{1}{c}{Worst-group} & \multicolumn{1}{c}{Avg Acc} \\
\midrule
  \alg & JTT & $86.2 \pm 3.6$ & $92.0 \pm 0.8$ \\
 \alg & GDRO(/\textsc{George}/EIIL)	& $87.6 \pm  0.8$ & $89.4 \pm  1.3$ \\
 \rowcolor{lightgray!40} \alg & \alg & $\mathbf{91.6 \pm 0.8}$ & $\mathbf{96.2 \pm 0.6}$\\
\bottomrule
    \end{tabular}%
    }
    \vspace{-3mm}
\end{table}

\subsection{Discovering Natural Spurious Correlations in Restricted ImageNet} 
Next, we show the applicability of \alg to discover and mitigate spurious correlations in Restricted ImageNet \citep{taghanaki2021robust}, a 9-superclass subset of ImageNet. Here, we train ResNet-50 from scratch. 
See \ref{sec:imagenet-app} for more details on the dataset and experiment.

\vspace{-1mm}
\paragraph{\alg Discovers Spurious Correlations
}
We observe \alg clusters during the initial training epochs. Inspecting the clusters with the highest fraction of misclassified examples to another class, we find that many frog images are misclassified as insects. \cref{fig:epoch8} shows examples from the two groups \alg finds for the Insect class at epoch 8, where clusters with spurious features are visually evident 
\footnote{Since the model is not pretrained, it is expected that the spurious clusters form slightly later. For pretrained models, spurious clusters form very early, as shown in \cref{tab:spurious}\looseness=-1}. 
GradCAM reveals an obvious spurious correlation between ``green leaf'' and the insect class that is maintained until the end of the training, as illustrated in \cref{fig:end}. We also observe a large gap between the confidence of examples in the two groups. This indicates that the model has learned the spurious feature early in training.\looseness=-1

\begin{table}
    \centering
    \caption{Discovering \& mitigating spurious correlations in Restricted ImageNet. \alg infers groups more accurately and improves both insect and frog minority accuracy by 1.2\% and 11.5\% respectively, with only a minor drop in total accuracy. Note that the group with worst-group accuracy changes during the training.}
    \resizebox{\columnwidth}{!}{%
    \begin{tabular}{cccccc}
    \toprule
    \multirow{2}{*}{Method} & Minority & Test & Insect & Frog & W-G \\
     & Recall &  Avg Acc & Min Acc & Min Acc & Acc \\
    \midrule
    ERM & - & $96.0\%$ & $91.7\%$ & $80.8\%$ & $80.8\%$ \\
    CB & - & $95.9\%$ & $\mathbf{93.7\% \uparrow}$ & $80.8\% -$ & $80.8\% -$  \\ 
    EIIL  & $78.8\%$ & $93.1\%$ & $88.3\% \downarrow $ & $69.2\% \downarrow$ & $69.2\% \downarrow$ \\
    JTT &  $82.6\%$ & $92.8\%$ & $75.0\% \downarrow$ & $\mathbf{92.3\% \uparrow}$ & $75.0\% \downarrow$ \\
    GEORGE & $85.4\%$ & $94.8\%$ & $89.4\% \downarrow$ & $80.8\% -$ & $80.8\% -$\\
    \rowcolor{lightgray!40} \textbf{\alg}  & $\mathbf{92.8\%}$ & $95.4\%$ & $\mathbf{92.9\% \uparrow}$ & $\mathbf{92.3\% \uparrow}$ & $\mathbf{92.3\% \uparrow}$ \\
    \bottomrule
    \end{tabular}
    }
    
    \label{tab:imagenet}
\vspace{-3mm}
\end{table}

\vspace{1mm}
\textbf{\alg Discovers Spurious Correlations without Relying on Group-labeled Validation Data.}  
State-of-the-art group inference baselines heavily rely on a group-labeled validation set to identify the time of group inference during training with ERM. 
While \alg can also benefit from a group-labeled validation, this is not essential. 
In fact, our theoretical results reveal that the range for inference time should be within the initial epochs. Thus we only visually inspect a few initial epochs (2, 4, 6, 8) to verify the spurious correlations.
This sets \alg apart as a more generally applicable method for discovering and mitigating spurious correlations, even in the absence of a group-labeled validation set. \looseness=-1

\vspace{1mm}
\textbf{\alg Achieves State-of-the-art Accuracy on Minority Groups.} Based on the spurious correlations \alg discovered, we manually labeled the background of both training and test data for the \textit{insect} and \textit{frog} classes, and these group labels to tune the baseline group inference methods.
\cref{tab:imagenet} shows
\alg separates the insect majority group with the spurious correlation better than other group inference methods and improves both insect and frog minority accuracy by 1.2\% and 11.5\% respectively, with only a minor drop in total accuracy. Note that group with worst-group accuracy changes during the training.
 CB only improves insect minority accuracy. JTT %
 decreases the model's accuracy on the insect minority a lot while improving the frog minority. EIIL decreases both minority and total accuracy as it finds the least majority. 
 Unlike the baselines, \alg effectively balances groups, mitigating spurious correlations. \looseness=-1

\begin{figure}[t]
  \begin{subfigure}[t]{0.23\textwidth}
         \centering
         \includegraphics[width=\textwidth]{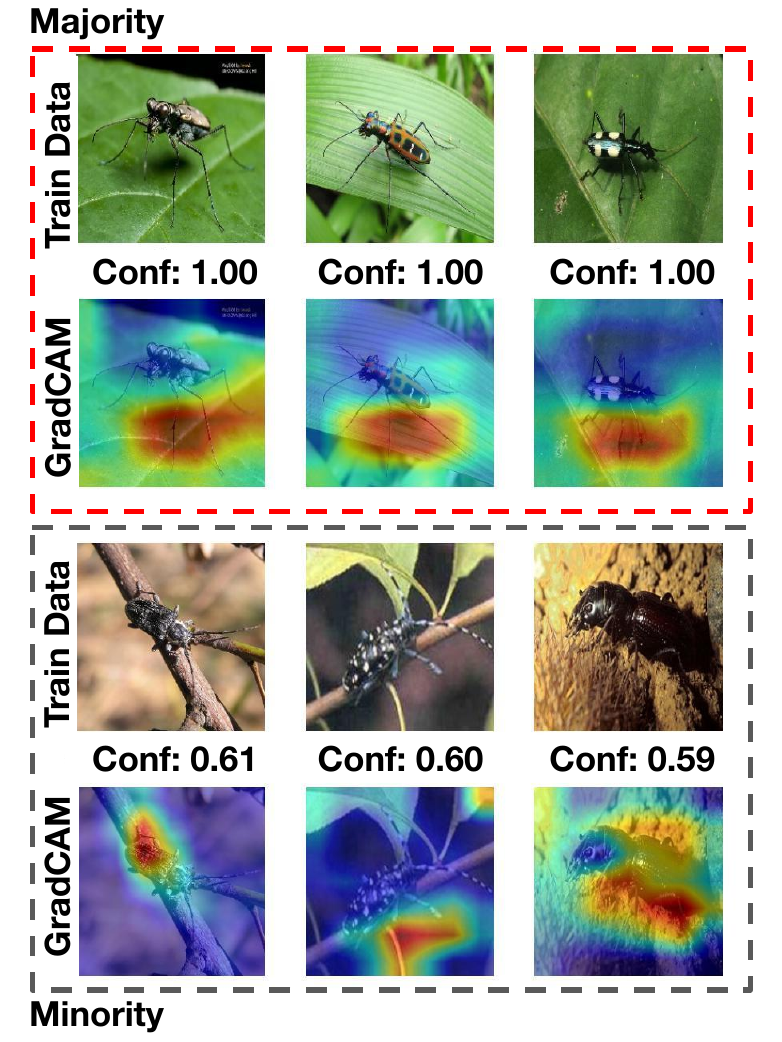}
         \caption{ERM epoch 8: model learns the spurious feature ``green leaf" in Insect class.}
         \label{fig:epoch8}
     \end{subfigure}
     \hfill
    \begin{subfigure}[t]{0.23\textwidth}
         \centering
         \includegraphics[width=\textwidth]{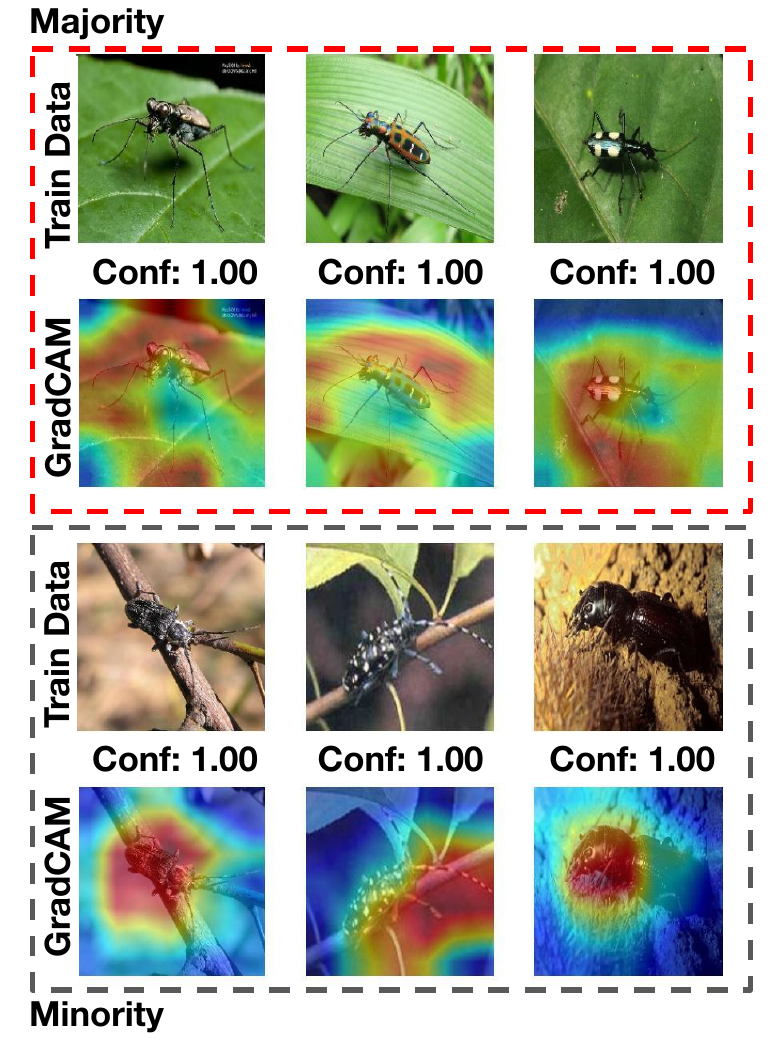}
         \caption{ERM end of training: model keep relying on ``green leaf", instead of Insect.}
         \label{fig:end}
     \end{subfigure}
  \caption{\alg-discovered spurious correlation between ``green leaf'' $\&$ ``insect'' in Restricted ImageNet. 
  \looseness=-1}
  \label{fig:imagenet}
\end{figure}

%% file: sections/5_conclusion.tex
\section{CONCLUSION}
We studied the effect of simplicity bias of SGD on learning spurious features.
Specifically, we analyzed a two-layer fully-connected neural network and showed that spurious features can be identified early in training based on model output. If spurious features have a low noise-to-signal ratio, they dominate the network's output, overshadowing core features. Based on the above theoretical insights, we proposed \alg, which separates majority and minority groups by clustering the model output early in training. Then, it applies importance sampling based on the cluster sizes to make the groups relatively balanced. It outperforms state-of-the-art methods in worst-group accuracy on benchmark datasets 
and scales well to large-scale applications.
Importantly, it can deal with multiple spurious correlations, 
minimizes the need for hyperparameter tuning, and can discover spurious correlations in realistic scenarios like Restricted ImageNet, early in training.

%% file: sections/6_appendix.tex
\section{SIMPLICITY BIAS} \label{apdx:simplicity_bias}
A recent body of work revealed that the neural network trained with (stochastic) gradient methods can be approximated on the training data by a {linear} function early in training \citep{hermann2020shapes,hu2020surprising,nakkiran2019sgd,neyshabur2014search,pezeshki2021gradient,shah2020pitfalls}. We hypothesize that a slightly stronger statement holds, namely the approximation still holds if we isolate a core or spurious feature from an example and input it to the model.
\begin{assumption} [simplicity bias on core and spurious features, informal] \label{assumption:simplicity_bias}
Suppose that $f^{lin}$ is a linear function that closely approximates $f(\vct{x}; \mtx{W}, \vct{z})$ on the training data. Then $f^{lin}$ also approximates $f$ on input either a core feature or a spurious feature corresponding to a majority group in some class, that is
\begin{align*} 
    f^{lin}(\vct{v}_c) &\approx f(\vct{v}_c; \mtx{W}, \vct{z}) \qquad \forall c \in \mathcal{C} \\
    f^{lin}(\vct{v}_s) &\approx f(\vct{v}_s; \mtx{W}, \vct{z}) \qquad \forall s \in \mathcal{A}
\end{align*}
\end{assumption}
Intuitively, every core feature and every spurious feature corresponding to a majority group is well represented in the training dataset, and since it is known that the linear model and the full neural network agree on the training dataset, we can expect them to agree on such features as well. Note that spurious features that do not appear in majority groups may not be well represented in the training dataset, hence we do not require that the linear model approximates the neural network well on such features.

\begin{figure}[!htb]
  \centering
    \begin{minipage}{.49\textwidth}
    \includegraphics[width=0.95\columnwidth]{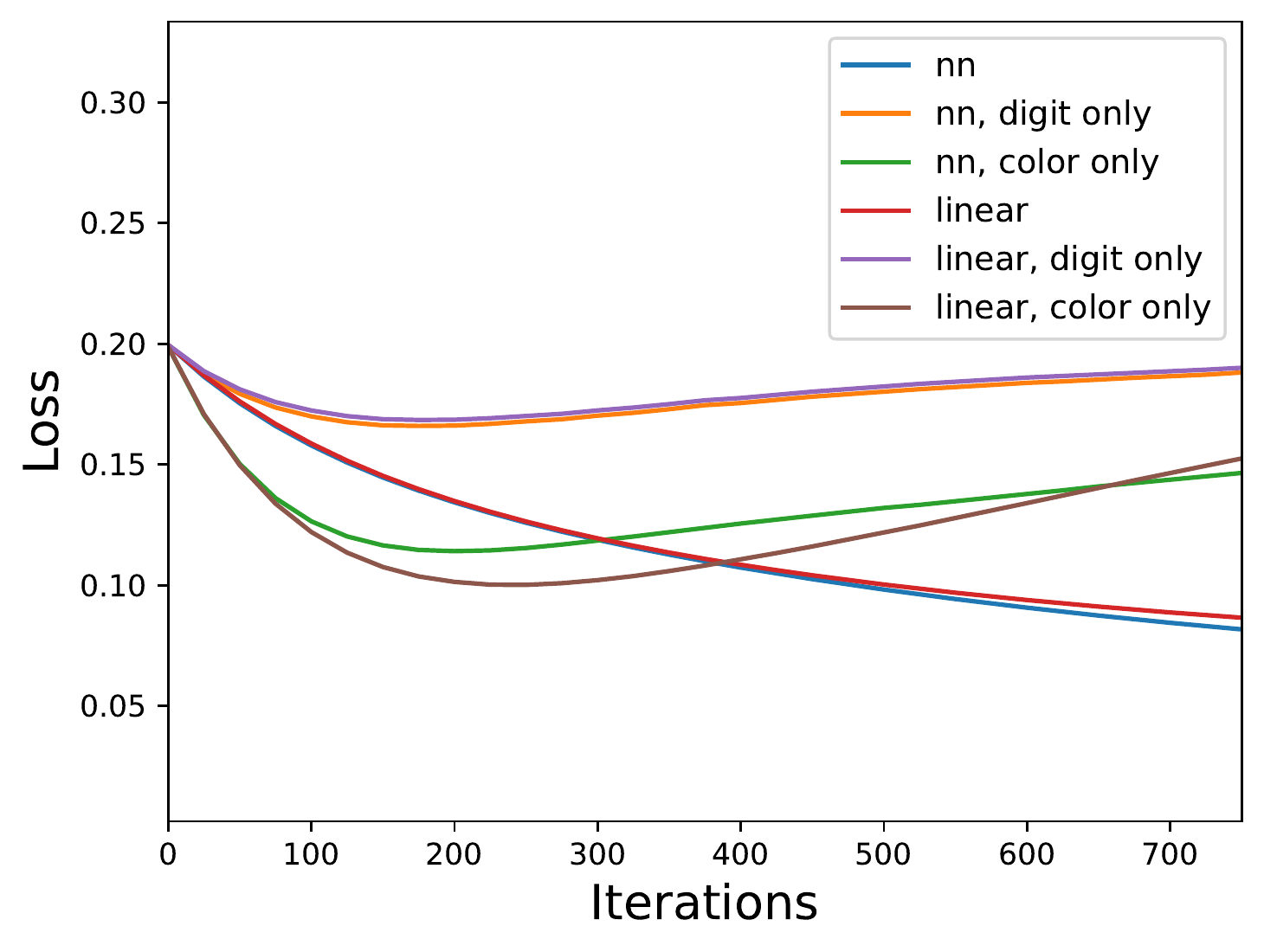}
    \caption{\footnotesize A comparison between the losses of a two-layer network and a simple linear model on the training set, spurious features (color only), and core feature (digit only).}
    \label{fig:rebuttal-assumption}
     \end{minipage}
     \hfill
    \begin{minipage}{.49\textwidth}
    \includegraphics[width= 0.95\columnwidth]{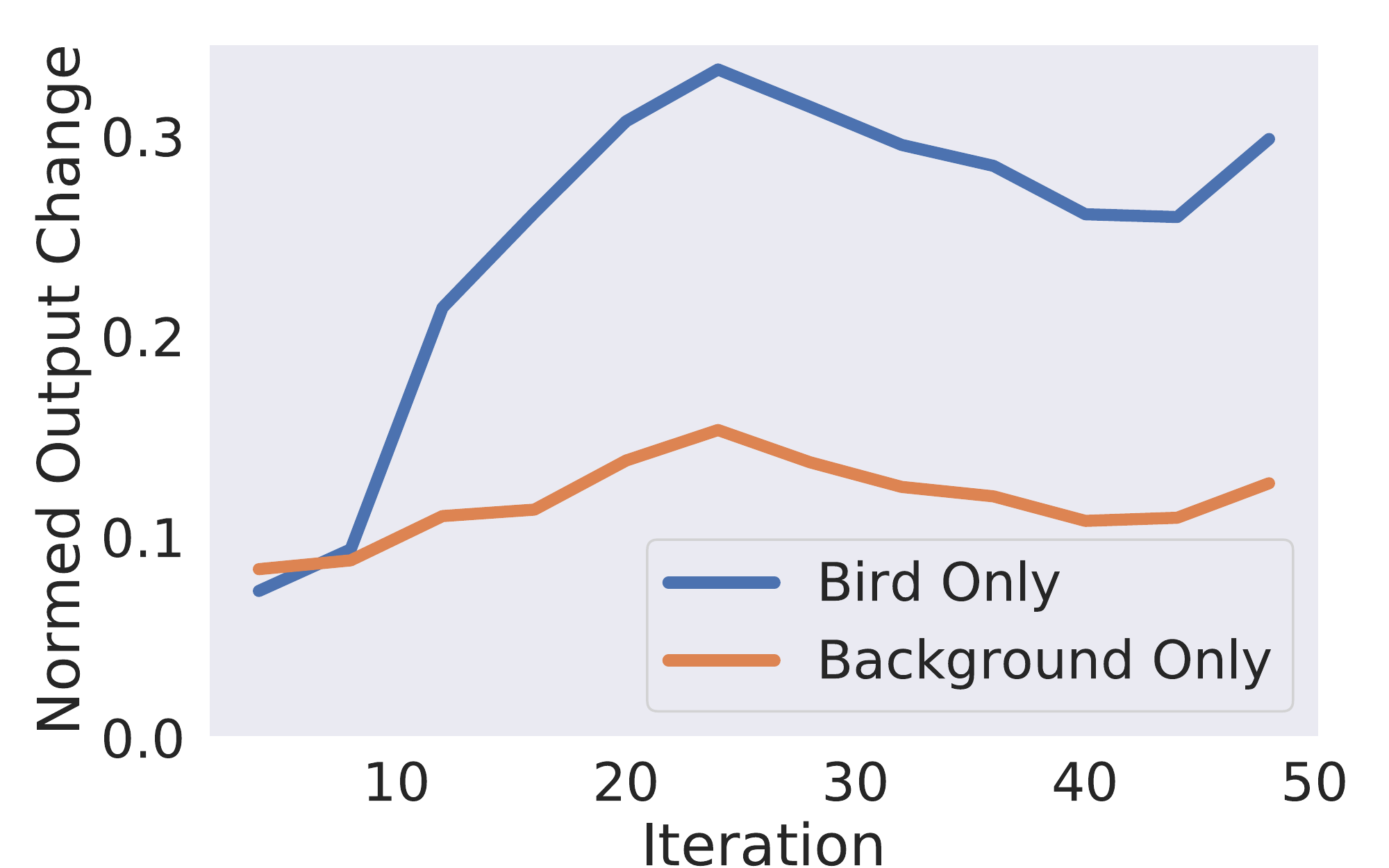}
    \caption{\footnotesize Replicate of Figure 1b on Waterbirds. Inputting only the background (orange line) does not change the model output much (indicating that the background is learned by the model) while inputting only the bird changes the output to a large extent (indicating that the bird is not learned by the model).}
    \label{fig:rebuttal-pred-diff}
    \end{minipage}
     
\end{figure}

Moreover, we verify \cref{assumption:simplicity_bias} empirically on CMNIST in \cref{fig:rebuttal-assumption}, which shows that a two-layer neural network and the approximating linear model are close even when isolating a core or spurious feature.

The formal statement is provided below as \cref{assumption:simplicity_bias_full}.

\paragraph{Setting} \label{apdx:setting}
We now introduce the formal mathematical setting for the theory. Let $\mathcal{D} = \{(\vct{x}_i, y_i)\}_{i=1}^n \subset \mathbb{R}^d \times \mathbb{R}$, be a dataset with covariance $\mtx{\Sigma}$. Define the data matrix $\mtx{X} = \begin{bmatrix} \vct{x}_1 & \dots & \vct{x}_n \end{bmatrix}^{\top}$ and the label vector $\vct{y} = \begin{bmatrix} y_1 & \dots & y_n \end{bmatrix}^{\top}$. 
We use $\|\cdot\|$ to refer to the Euclidean norm of a vector or the spectral norm of the data.

Following \cite{hu2020surprising}, we make the following assumptions:

\begin{assumption}[input distribution]
The data has the following properties (with high probability): 
\begin{align*}
    \frac{\|\vct{x}_i\|^2}{d} &= 1 \pm \OO(\sqrt{\frac{\log{n}}{d}}),  \forall i \in [n] \\
    \frac{|\inner{\vct{x}_i}{\vct{x}_j}|}{d} &=   \OO(\sqrt{\frac{\log{n}}{d}}),  \forall i, j \in [n], i \neq j \\
    \|\mtx{X} \mtx{X}^{\top} \| &= \Theta(n)
\end{align*}
\end{assumption}

\begin{assumption} [activation function]
The activation $\phi(\cdot)$ satisfies either of the following:
\begin{itemize}
    \item smooth activation: $\phi$ has bounded first and second derivative
    \item piecewise linear activation: 
    $$\phi(z) = \begin{cases}
        z & z \geq 0 \\
        az & z < 0
    \end{cases}$$
\end{itemize}
\end{assumption}

\begin{assumption} [initialization]
The weights $(\mtx{W}, \vct{v})$ are initialized using symmetric initialization:
\begin{align*}
    \vct{w}_1, \dots, \vct{w}_{\frac{m}{2}} \sim \mathcal{N}(\vct{0}_d,\mtx{I}_d), \qquad \vct{w}_{i + \frac{m}{2}} = \vct{w}_i (\forall i \in 1, \dots, \frac{m}{2}) \\
    v_1, \dots, v_{\frac{m}{2}} \sim \text{Unif}(\{-1, 1\}), \qquad v_{i + \frac{m}{2}} = -v_i (\forall i \in 1, \dots, \frac{m}{2})
\end{align*}
\end{assumption}

It is not hard to check that the concrete scenario we choose in our analysis satisfies the above assumptions. Now, given the following assumptions, we leverage the result of \citep{hu2020surprising}:

\begin{theorem}[\citep{hu2020surprising}]\label{thm:hu}
Let $\alpha\!\in\!(0,1/4)$ be a fixed constant. Suppose $d$ is the input dimensionality, $\frac{\inner{\vct{x}_i}{\vct{x}_j}}{d} =  \mathbb{1}_{i=j} \pm  \OO(\sqrt{\frac{\log{n}}{d}}),  \forall i, j \in [n]$, the data matrix $\X\!=\!\{\x_i\}_{i=1}^n$ has spectral norm $\norm{\X \X^{\top}}=\Theta(n)$, and for the labels we have $|y_i|\leq 1~\forall y_i$. Assume the number of training samples $n$ and the network width $m$ satisfy $n,m = \Omega(d^{1+\alpha}), n,m\leq d^{\mathcal{O}(1)}$, and the learning rate $\eta\ll d$. 
Then, there exist a universal constant $C$, such that with high probability for all $0\leq t\leq T=C\cdot\frac{d\log d}{\eta}$, the network $f(\w_t,\X)$ {trained with GD} is very close to a linear function $f^{lin}(\bet,\X)$: \looseness=-1
\begin{align}\label{eq:linear}
    &\frac{1}{n}\sum_{i=1}^n (
    f^{lin}(\bet_t,\X) \!-\! f(\w_t,\X))^2 
    \leq 
    \frac{\eta^2t^2}{d^{2 + \Omega(\alpha)}} \leq \frac{1}{d^{\Omega(\alpha)}}. %
\end{align} 
\end{theorem}

In particular, the linear model $f^{lin}(\bet,\X)$ operates on the transformed data $\vct{\psi}(\vct{x})$, where
\begin{align*}
    \vct{\psi}(\vct{x}) &= \begin{bmatrix} \sqrt{\frac{2}{d}} \zeta \vct{x} \\ \sqrt{\frac{3}{2d}} \nu \\ \vartheta_0 + \vartheta_1 (\frac{\| \vct{x} \|}{\sqrt{d}} - 1) + \vartheta_2 (\frac{\| \vct{x} \|}{\sqrt{d}} - 1)^2 \end{bmatrix} \\
    \zeta &= \E_{g \sim \mathcal{N}(0, 1)}[\phi'(g)] \\
    \nu &= \E_{g \sim \mathcal{N}(0, 1)}[g\phi'(g)] \sqrt{\frac{\Tr[\mtx{\Sigma}^2]}{d}} \\
    \vartheta_0 &= \E_{g \sim \mathcal{N}(0, 1)}[\phi(g)] \\
    \vartheta_1 &= \E_{g \sim \mathcal{N}(0, 1)}[g\phi'(g)] \\
    \vartheta_2 &= \E_{g \sim \mathcal{N}(0, 1)}[(\frac{1}{2}g^3 - g)\phi'(g)]
\end{align*}
Note that $\vct{\psi}(\vct{x})$ consists of a scaled version of the data, a bias term, and a term that depends on the norm of the example. We will adopt the notation $f^{lin}(\vct{x}; \vct{\beta}) = \vct{\psi}(\vct{x})^{\top} \vct{\beta}$ for the linear model.

We can now formally state \cref{assumption:simplicity_bias}:
\begin{assumption} [formal version of \cref{assumption:simplicity_bias}] \label{assumption:simplicity_bias_full} 
Suppose that \cref{thm:hu} holds. Then with high probability, for all such $t$ the following also holds for all $c \in \mathcal{C}$ and for all $s \in \mathcal{A}$:
\begin{align*}
    |f^{lin}(\bet_t,\vct{v}_c) \!-\! f(\w_t,\vct{v}_c)| \leq  \OO(\frac{\eta t}{d^{1 + \Omega(\alpha)}}), \\
    |f^{lin}(\bet_t,\vct{v}_s) \!-\! f(\w_t,\vct{v}_s)| \leq  \OO(\frac{\eta t}{d^{1 + \Omega(\alpha)}}).
\end{align*}
\end{assumption}

We will assume the former holds in the proof of the following theorems, although as we will see the assumption is unnecessary for \cref{thm:cluster}.

\section{PROOF FOR THEOREMS}
\subsection{Notation}
For the analysis, we split $\vct{\beta}$ into its components corresponding to the data, bias and norm parts of $\vct{\psi}$; that is $\vct{\beta} = \begin{pmatrix} \vct{\beta}' \\ \beta_{bias} \\ \beta_{norm} \end{pmatrix}$ for $\vct{\beta}' \in \mathbb{R}^d, \beta_{bias} \in \mathbb{R}, \beta_{norm} \in \mathbb{R}$. We use the inner product between $\vct{\beta}'$ and a feature $\vct{v}$ to understand how well the linear model learns a feature $\vct{v} \in \mathbb{R}^d$. With slight abuse of notation, we will simply write $\inner{\vct{\beta}}{\vct{v}}$ to mean $\inner{\vct{\beta}'}{\vct{v}}$.

We also define the matrix $\mtx{\Phi} = \begin{bmatrix} \vct{\phi}_1 & \dots & \vct{\phi}_n \end{bmatrix}^{\top}$.

\subsection{Proof of \cref{thm:lin_param} and \cref{thm:cluster}} \label{apdx:proof_early_dynamics}
\linparam*
\cluster*

As in \cite{hu2020surprising}, we will conduct our analysis under the high probability events that $\|\mtx{\Psi}^{\top} \mtx{\Psi}\| =  \OO(\frac{n}{d})$ and for all training data $\vct{x}, \frac{\| \vct{x} \|}{\sqrt{d}} = 1 \pm  \OO(\sqrt{\frac{\log{n}}{d}})$.

Starting from the rule of gradient descent
\begin{align*}
    \vct{\beta}(t+1) &= \vct{\beta}(t) - \frac{\eta}{n}\mtx{\Psi}^{\top}(\mtx{\Psi}\vct{\beta}(t) - \vct{y}) \\
    &= \left(I - \frac{\eta}{n}\mtx{\Psi}^{\top}\mtx{\Psi}\right)\vct{\beta}(t) + \frac{\eta}{n} \mtx{\Psi}^{\top} \vct{y}
\end{align*}
Let $\mtx{A} = \mtx{I} - \frac{\eta}{n}\mtx{\Psi}^{\top}\mtx{\Psi}, \vct{b} = \frac{\eta}{n}\mtx{\Psi}^{\top} \vct{y}$. Also, $\mtx{A}$ can be diagonalized as $\mtx{A} = \mtx{V}\mtx{D}\mtx{V}^{\top}$. Since $\|\mtx{\Psi}^{\top} \mtx{\Psi}\| =  \OO(\frac{n}{d})$, the eigenvalues of $\mtx{A}$, call them $\lambda_1, \dots, \lambda_d$, are of order $1 -  \OO(\frac{\eta}{d})$. For $t \geq 1$, the previous recurrence relation admits the solution
\begin{align*}
    \vct{\beta}(t) &= (\mtx{I} + \mtx{A} + \dots + \mtx{A}^{t-1})\vct{b} \\
    &= \mtx{V}(\mtx{I} + \mtx{D} + \dots + \mtx{D}^{t-1})\vct{V}^{\top}\vct{b}
\end{align*}
The eigenvalues of $\mtx{I} + \mtx{D} + \dots + \mtx{D}^{t-1}$ {is a geometric series $\frac{1 - \lambda_i^t}{1 - \lambda_i}$, where $\lambda_i$ are the eigenvalues of $D$. By the binomial theorem,} 
\begin{align*}
    1 + \lambda_i + \dots + \lambda_i^{t-1} &= \frac{1 - \lambda_i^t}{1 - \lambda_i} \\
    &= \frac{1 - (1 - t(1- \lambda_i) +  \OO(t^2(1-\lambda_i)^2))}{1 - \lambda_i}\\
    &= t +  \OO(t^2(1 - \lambda)) %
\end{align*}
When $t =  \OO(\sqrt{\frac{d^{1-\alpha}}{\eta}})$, the expression simplifies to $t + \OO(d^{-\alpha/2})$.
Thus we can approximate $\mtx{I} + \mtx{D} + \dots + \mtx{D}^{t-1} = t\mtx{I} + \mtx{\Delta}$, where $\|\mtx{\Delta}\| =  \OO(d^{-\frac{\alpha}{2}})$. Then
\begin{align*}
    \beta(t) = \mtx{V} (t \mtx{I} + \mtx{\Delta}) \mtx{V}^{\top} \vct{b} = t \vct{b} + \mtx{\Delta}_1\vct{b} 
\end{align*}
where $\mtx{\Delta}_1 = \mtx{V}\mtx{\Delta}\mtx{V}^{\top}$ also satisfies $\|\mtx{\Delta}\| = \OO(d^{-\frac{\alpha}{2}})$.

From here we may calculate the following: the alignment of $\vct{\beta}$ with a core feature $\vct{v}_{c}$ is
\begin{align} \label{core_feature_alignment}
    \inner{\vct{v}_{c}}{\vct{\beta}} &= \inner{\vct{v}_c}{\frac{\eta t}{n}\mtx{\Psi}^{\top} \vct{y} + \mtx{\Delta}_1 \frac{\eta}{n}\mtx{\Psi}^{\top} \vct{y}} \\
    &= \frac{\eta}{n} \sum_{i=1}^n \inner{\vct{v}_c}{t y_i \vct{\psi}_i + \mtx{\Delta}_1 y_i \vct{\psi}_i} \\
    &= \sqrt{\frac{2}{d}} \frac{\eta \zeta c \|\vct{v}_c\|}{n}(t \pm \OO(d^{-\frac{\alpha}{2}}))(\|\vct{v}_c\| n_c \pm \OO(\sigma_{c} \sqrt{n})) \\
    &= \sqrt{\frac{2}{d}} \eta \zeta c \|\vct{v}_c\|^2 t \left(\frac{n_c}{n} \pm \OO(d^{-\Omega(\alpha)})\right)
\end{align}
and the alignment with a spurious feature $\vct{v}_s$ is
\begin{align} \label{spurious_feature_alignment}
    \inner{\vct{v}_s}{\vct{\beta}} &= \inner{\vct{v}_s}{\frac{\eta t}{n}\Psi^{\top} \vct{y} + \mtx{\Delta}_1 \frac{\eta}{n}\Psi^{\top} \vct{y}} \\
    &= \frac{\eta}{n} \sum_{i=1}^n \inner{\vct{v}_s}{t y_i \vct{\psi}_i + \mtx{\Delta}_1 y_i \vct{\psi}_i} \\
    &= \sqrt{\frac{2}{d}} \frac{\eta \zeta c \|\vct{v}_s\|}{n} (t \pm \OO(d^{-\frac{\alpha}{2}})) (\|\vct{v}_s\| (n_{c,s} - n_{c',s}) \pm \OO(\sigma_{s} \sqrt{n})) \\
    &= \sqrt{\frac{2}{d}} \eta \zeta c \|\vct{v}_s\|^2 t \left(\frac{n_{c,s} - n_{c',s}}{n} \pm \OO(d^{-\Omega(\alpha)})\right) 
\end{align}
{Eq. \ref{core_feature_alignment}, \ref{spurious_feature_alignment} hold by
substituting $\beta = \frac{\eta t}{n}\Psi^{\top} \vct{y} + \mtx{\Delta}_1 \vct{b}$ error derived earlier and considering the inner product of every column of $\Psi^{\top}$ with the core/spurious feature.} The effect of the noise is captured by the $\OO(\sigma \sqrt{n})$ terms, following standard concentration inequalities, and we used the fact that $\frac{1}{\sqrt{n}} = \OO(d^{-\Omega(\alpha)})$. 

In addition, we calculate that 
\begin{align*}
    \beta_{norm}(t) = (tI + \mtx{\Delta}_1) \sum_{i=1}^n y_i\left(\vartheta_0 + \vartheta_1 (\frac{\| \vct{x}_i \|}{\sqrt{d}} - 1) + \vartheta_2 (\frac{\| \vct{x}_i \|}{\sqrt{d}} - 1)^2\right) = \OO(\frac{\eta t}{\sqrt{n}}) 
\end{align*}

Now the result transfers to the full neural network under \cref{assumption:simplicity_bias_full}, namely 
\begin{align} 
    f(\vb_c;\W_t,\z_t) &= f^{lin}(\bet_t,\vct{v}_c) \pm \OO(\frac{\eta t}{d^{1 + \Omega(\alpha)}}) \\
    &= \frac{2\eta \zeta^2 c \|\vct{v}_c\|^2 t}{d}  \left(\frac{n_c}{n} \pm \OO(d^{-\Omega(\alpha)})\right), \\
   f(\vb_s;\W_t,\z_t) \!&=\! f^{lin}(\bet_t,\vct{v}_s) \pm \OO(\frac{\eta t}{d^{1 + \Omega(\alpha)}}) \\ 
   &= \frac{2 \eta \zeta^2 c \|\vct{v}_s\|^2 t}{d} \left(\frac{n_{c,s} - n_{c',s}}{n} \pm \OO(d^{-\Omega(\alpha)})\right) ,
\end{align}
This proves \cref{thm:lin_param}.

Then for the predictions at time $t$ for an example in class $c = 1$, group $g_{1,s}$: 
\begin{align*}
    \vct{\psi}(\vct{x})^{\top} \vct{\beta}(t) &= \sqrt{\frac{2}{d}} \zeta \vct{x}^{\top} \vct{\beta}' + \sqrt{\frac{3}{2d}} \nu \beta_{bias}(t) + \beta_{norm}(t) \left(\vartheta_0 + \vartheta_1 (\frac{\| \vct{x} \|}{\sqrt{d}} - 1) + \vartheta_2 (\frac{\| \vct{x} \|}{\sqrt{d}} - 1)^2 \right) \\
    &= \sqrt{\frac{2}{d}} \zeta (\vct{v}_1 + \vct{v}_{s} + \vct{\xi})^{\top} \vct{\beta}' + \sqrt{\frac{3}{2d}} \nu \beta_{bias}(t) + \vartheta_0  \beta_{norm}(t) \pm O\left(\eta t \sqrt{\frac{\log{n}}{nd}}\right)
\end{align*}
 
We have a few cases 
\begin{enumerate} \label{prediction_cases}
    \item $g_{1,k}$ is a majority group. In this case
    \begin{align*}
        \vct{\psi}(\vct{x})^{\top} \vct{\beta}(t) &\geq \frac{2\eta \zeta^2 t}{d} \left(\frac{n_1 \|\vct{v}_c\|^2}{n} + \frac{\|\vct{v}_s\|^2 (K - k)}{n}  + \inner{\vct{\xi}}{\frac{1}{n}\mtx{X}^{\top}\vct{y}} \pm \OO(d^{-\Omega(\alpha)})\right)  \\ & \quad+ \sqrt{\frac{3}{2d}} \nu \beta_{bias}(t) + \vartheta_0  \beta_{norm}(t) \pm O\left(\eta t \sqrt{\frac{\log{n}}{nd}}\right)
    \end{align*}
    \item $g_{1,k}$ is a minority group and $g_{-1,k}$ is a majority group. In this case
    \begin{align*}
        \vct{\psi}(\vct{x})^{\top} \vct{\beta}(t) &\leq \frac{2\eta \zeta^2 t}{d} \left(\frac{n_1 \|\vct{v}_c\|^2}{n} - \frac{\|\vct{v}_s\|^2 (K - k)}{n}  + \inner{\vct{\xi}}{\frac{1}{n}\mtx{X}^{\top}\vct{y}} \pm \OO(d^{-\Omega(\alpha)})\right) \\ & \quad + \sqrt{\frac{3}{2d}} \nu \beta_{bias}(t) + \vartheta_0  \beta_{norm}(t) \pm O\left(\eta t \sqrt{\frac{\log{n}}{nd}}\right)
    \end{align*}

    \item $g_{1,k}$ is such that no majority groups have the spurious feature. In this case
    \begin{align*}
        \vct{\psi}(\vct{x})^{\top} \vct{\beta}(t) &= \frac{2\eta \zeta^2 t}{d} \left(\frac{n_1 \|\vct{v}_c\|^2}{n} + \frac{\|\vct{v}_s\|^2 \tilde{k}}{n} + \inner{\vct{\xi}}{\frac{1}{n}\mtx{X}^{\top}\vct{y}} \pm \OO(d^{-\Omega(\alpha)})\right) \\ & \quad + \sqrt{\frac{3}{2d}} \nu \beta_{bias}(t) + \vartheta_0  \beta_{norm}(t) \pm O\left(\eta t \sqrt{\frac{\log{n}}{nd}}\right), \qquad |\tilde{k}| \leq k
    \end{align*}
\end{enumerate}
Now 
\begin{align}
    \inner{\vct{\xi}}{\frac{1}{n}\mtx{X}^{\top}\vct{y}} &= \sum_{c \in \{\pm 1\}} \frac{\|\vct{v}_c\| n_c}{n} \inner{\vct{\xi}}{\vct{v}_c} + \sum_{s} \frac{\|\vct{v}_s\|(n_{1,s}-n_{-1,s})}{n} \inner{\vct{\xi}}{\vct{v}_s} + \inner{\vct{\xi}}{\frac{1}{n}\sum_{i=1}^n \vct{\xi}_i y_i} \\ 
    &= \sum_{c \in \{\pm 1\}} \frac{\|\vct{v}_c\|n_c}{n} \inner{\vct{\xi}}{\vct{v}_c} + \sum_{s} \frac{\|\vct{v}_s\|(n_{1,s}-n_{-1,s})}{n} \inner{\vct{\xi}}{\vct{v}_s} \pm O\left(\sqrt{\frac{d}{n}}\right) \\
    &\sim \mathcal{N}(0, \kappa) \pm \OO(d^{-\Omega(\alpha)})
\end{align}
Finally, observe that $O\left(\eta t \sqrt{\frac{\log{n}}{nd}}\right) = \OO(d^{-1 - \Omega(\alpha)})$. Combining all these results and setting $\rho_1 = \frac{2\eta \zeta^2 c t}{d}, \rho_2 = \frac{\rho_1 n_1 \|\vct{v}_c\|^2}{n} + \sqrt{\frac{3}{2d}} \nu \beta_{bias}(t) + \vartheta_0  \beta_{norm}(t)$ shows \cref{thm:cluster} when looking at the prediction of the linear model. Recall that \cite{hu2020surprising} showed that the average squared error in predictions between the linear model and the full neural network is $\OO(\frac{\eta^2 t^2}{d^{2+\Omega(\alpha)}})$. Then by Markov's inequality, we can guarantee that the predictions of the linear model differ by at most $\OO(\frac{\eta t}{d^{1+\Omega(\alpha)}})$ for at least $1 - \OO(d^{-\Omega(\alpha)})$ proportion of the examples. This error can be factored into the existing error term. Hence the result holds for the full neural network.

We can apply the same argument for the class $c'$. Thus \cref{thm:cluster} is proven.

Notably, \cref{thm:cluster} only depends on the closeness of the neural network and the initial linear model on the training data, hence does not rely on \cref{assumption:simplicity_bias_full}.

\subsection{Proof of \cref{thm:optimal_lin_param}}
\label{apdx:proof_optimal_lin_param}

\optimallinparam*

Let $g_{maj}$ be the total number of majority groups among all classes. Note that by the definition of majority groups, $g_{maj}$ is at most the number of classes, namely $2$ in the given analysis.

Since the classes are balanced with labels $\pm 1$, it is not hard to see that the bias term in the weights will always be zero, hence we may as well assume that we do not have the bias term. Abusing notation, we will still denote quantities by the same symbol, even though now the bias term has been removed.

First consider a model $\tilde{f} = \vct{\psi}^{\top} \tilde{\vct{\beta}}$ trained on the dataset $\mathcal{D}_{\maj}$, which only contains examples from the majority groups, and the norm term of $\psi$ is only the constant term $\vartheta_0$ instead of $\vartheta_0 + \vartheta_1 (\frac{\| \vct{x} \|}{\sqrt{d}} - 1) + \vartheta_2 (\frac{\| \vct{x} \|}{\sqrt{d}} - 1)^2$. Further, assume $\mathcal{D}_{\maj}$ has infinitely many examples so that the noise perfectly matches the underlying distribution. We prove the results in this simplified setting then extend the result using matrix perturbations.

We have
\begin{align*}
    \mathcal{L} &= \frac{1}{2}\E_{\mathcal{D}_{\maj}}[(\vct{\psi}_i^{\top} \tilde{\vct{\beta}} - y_i)^2] \\
    \nabla \mathcal{L} &= \E_{\mathcal{D}_{\maj}}[(\vct{\psi}_i^{\top} \tilde{\vct{\beta}} - y_i)\vct{\psi}_i]
\end{align*}
and the optimal $\tilde{\vct{\beta}}_{\ast}$ satisfies
\begin{align*}
    \tilde{\vct{\beta}}_{\ast} = \left(\E_{\mathcal{D}_{\maj}}[\vct{\psi}_i \vct{\psi}_i^{\top}]\right)^{\dagger} \E_{\mathcal{D}_{\maj}}[y_i \vct{\psi}_i]
\end{align*}
where $\dagger$ represents the Moore-Penrose pseudo-inverse.

Since the noise is symmetrical with respect to the classes, the bias and constant norm terms of $\vct{\beta}$ must be zero Formally, the first order condition implies
\begin{align*}
    \E_{\mathcal{D}_{\maj}}[\vct{\psi}_i \vct{\psi}_i^{\top}] \tilde{\vct{\beta}}_{\ast} &= \E_{\mathcal{D}_{\maj}}[y_i\vct{\psi}_i]
\end{align*}
As shorthand, denote $\alpha = 
    \begin{bmatrix}
        \psi_{bias} \\ \vartheta_0
    \end{bmatrix}^{\top}
    \begin{bmatrix}
        \tilde{\beta}_{bias}^{\ast} \\ \tilde{\beta}_{norm}^{\ast}
    \end{bmatrix}$
Decomposing the above vector equation based on the bias and constant norm terms versus the others, we have
\begin{align}
    \E_{\mathcal{D}_{\maj}} \left[\begin{bmatrix}
        \psi_{bias} \\ \vartheta_0
    \end{bmatrix} \left(\sqrt{\frac{2}{d}}\zeta \vct{x}_i \right)^{\top}
    \tilde{\vct{\beta}}' \right] +
    \begin{bmatrix}
        \psi_{bias} \\ \vartheta_0
    \end{bmatrix} 
    \alpha &= 
    \begin{bmatrix}
        0 \\ 0
    \end{bmatrix} \\
    \E_{\mathcal{D}_{\maj}} \left[\left(\sqrt{\frac{2}{d}}\zeta \vct{x}_i \right)^{\top} \tilde{\vct{\beta}}' \right] + \alpha &= 0
    \label{eq:lin_param_fonc_constant}
\end{align}
The remaining coordinates give the equation
\begin{align}
    \E_{\mathcal{D}_{\maj}} \left[\left(\sqrt{\frac{2}{d}}\zeta \vct{x}_i \right) \left(\sqrt{\frac{2}{d}}\zeta \vct{x}_i \right)^{\top} \tilde{\vct{\beta}}' \right] 
    + \E_{\mathcal{D}_{\maj}} \left[\left(\sqrt{\frac{2}{d}}\zeta \vct{x}_i \right) \alpha\right] &=  \E_{\mathcal{D}_{\maj}}\left[y_i\sqrt{\frac{2}{d}} \zeta \vct{x}_i \right]
    \label{eq:lin_param_fonc_feature}
\end{align}
Set $\vct{z} = \sum_{c} \frac{\vct{v}_c}{\|v_c\|^2}$. Observe that for all $\vct{x}_i$,
\begin{align*}
    \vct{z}^{\top} \vct{x}_i = 1 + \vct{z}^{\top} \vct{\xi}_i
\end{align*}
Taking the inner product of the latter equation \ref{eq:lin_param_fonc_feature} with $\vct{z}$ gives
\begin{align*}
    \E_{\mathcal{D}_{\maj}} \left[\sqrt{\frac{2}{d}}\zeta(1 + \vct{z}^{\top} \vct{\xi}_i) \left(\sqrt{\frac{2}{d}}\zeta \vct{x}_i \right)^{\top} \tilde{\vct{\beta}}' \right] 
    + \E_{\mathcal{D}_{\maj}} \left[\sqrt{\frac{2}{d}}\zeta(1 + \vct{z}^{\top} \vct{\xi}_i) \alpha \right] &=  \E_{\mathcal{D}_{\maj}}\left[y_i\sqrt{\frac{2}{d}} \zeta(1 + \vct{z}^{\top} \vct{\xi}_i) \right] \\
    \E_{\mathcal{D}_{\maj}} \left[\frac{2}{d}\zeta^2 (\vct{x}_i^{\top} \tilde{\vct{\beta}}' + \vct{z}^{\top} \vct{\xi}_i \vct{\xi}_i ^{\top} \tilde{\vct{\beta}}') \right] +\sqrt{\frac{2}{d}}\zeta\alpha &= 0 \\
    \E_{\mathcal{D}_{\maj}} \left[\frac{2}{d}\zeta^2 \vct{x}_i^{\top} \tilde{\vct{\beta}}'\right]  + \frac{2}{d}\zeta^2 \sum_c \frac{\sigma_c^2}{\|v_c\|^2} \vct{v}_c^{\top} \tilde{\vct{\beta}}' + \sqrt{\frac{2}{d}}\zeta\alpha &= 0 \\
    \E_{\mathcal{D}_{\maj}} \left[\sqrt{\frac{2}{d}}\zeta \vct{x}_i^{\top} \tilde{\vct{\beta}}'\right]  + \sqrt{\frac{2}{d}}\zeta R_c^2 \sum_c \vct{v}_c^{\top} \tilde{\vct{\beta}}' + \alpha &= 0
\end{align*}
Combined with equation \ref{eq:lin_param_fonc_constant}, we conclude that $\sum_c \vct{v}_c^{\top} \tilde{\vct{\beta}}' = 0$. But since the classes are balanced, this implies that $\E_{\mathcal{D}_{\maj}}[\vct{v}_{c_i}^{\top} \tilde{\vct{\beta}}'] = 0$. A similar argument shows that $\E_{\mathcal{D}_{\maj}}[\vct{v}_{s_i}^{\top} \tilde{\vct{\beta}}'] = 0$. We conclude that $\E_{\mathcal{D}_{\maj}}[\vct{x}_{i}^{\top} \tilde{\vct{\beta}}'] = 0$, hence $\alpha = 0$. Now since the solution $\tilde{\vct{\beta}}^{\ast}$ lies in the span of the data, we must have
\begin{align*}
   \left\|\begin{bmatrix}
        \tilde{\beta}_{bias}^{\ast} \\ \tilde{\beta}_{norm}^{\ast}
    \end{bmatrix}\right\|^2 = \alpha  = 0
\end{align*}
We conclude that $\tilde{\beta}_{bias}^{\ast} = \tilde{\beta}_{norm}^{\ast} = \alpha = 0$, as claimed.

Thus the loss becomes
\begin{align}
    \mathcal{L} &= \frac{1}{2}\E_{(\vct{x}_i, y_i) \sim \mathcal{D}_\maj}\left[(\sqrt{\frac{2}{d}}\zeta \vct{x}_i^{\top} \tilde{\vct{\beta}}' - y_i)^2\right] \\
    &= \frac{1}{2}\E_{\mathcal{D}_{\maj}}\left[(\sqrt{\frac{2}{d}}\zeta (\vct{v}_{c_i} + \vct{v}_{s_i} + \vct{\xi}_i)^{\top} \tilde{\vct{\beta}}' - y_i)^2\right] \\
    &= \frac{1}{2}\E_{\mathcal{D}_{\maj}}\left[(\sqrt{\frac{2}{d}}\zeta (\vct{v}_{c_i} + \vct{v}_{s_i})^{\top} \tilde{\vct{\beta}}' - y_i)^2 + (\sqrt{\frac{2}{d}}\zeta \vct{\xi}_{i}^{\top} \tilde{\vct{\beta}}')^2\right] \\
    \label{apdx:decomposed_loss}
    &= \frac{1}{2}\E_{\mathcal{D}_{\maj}}\left[(\sqrt{\frac{2}{d}}\zeta (\vct{v}_{c_i} + \vct{v}_{s_i})^{\top} \tilde{\vct{\beta}}' - y_i)^2 \right] + \frac{\zeta^2}{d} \tilde{\vct{\beta}}'^{\top} \mtx{\Sigma}_{\xi} \tilde{\vct{\beta}}'
\end{align}

Consider the model $\vct{\beta}_s$ which only learns the spurious features of majority groups
\begin{align*}
    \vct{\beta}_s' &= \sqrt{\frac{d}{2}} \frac{1}{\zeta} \sum_{g_{c,s} \text{is a majority group}} \frac{c \vct{v}_s}{\|\vct{v}_s\|^2}.
\end{align*}

Note that for any example in a majority group, $(\vct{v}_{c_i} + \vct{v}_{s_i} )^{\top} \vct{\beta}_s' - y_i = 0$. Thus
\begin{align*}
    \mathcal{L} &= \frac{\zeta^2}{d} \tilde{\vct{\beta}}'^{\top} \mtx{\Sigma}_{\xi} \tilde{\vct{\beta}}' \\
    &= \sum_{\vct{v}_s \text{is spurious}} \frac{\sigma_s^2}{2\|\vct{v}_s\|^2}\\ 
    &\leq \frac{g_{\maj} R^2}{2}
\end{align*}

The loss for the optimal model must be smaller. But the loss due to the last term in \cref{apdx:decomposed_loss} along a core feature alone is 
\begin{align*}
    \frac{\zeta^2 \sigma_c^2}{\|\vct{v}_c\|^2 d}  \inner{\vct{v}_{c}}{\vct{\beta}_{\ast}'}^2 \leq  \frac{g_{\maj} R_s^2}{2}
\end{align*}
Rearranging gives 
\begin{align} \label{apdx:optimal_core_alignment}
    \inner{\vct{v}_c}{\vct{\beta}_{\ast}'}^2 &\leq  \frac{d g_{\maj} R^2 \|\vct{v}_c\|^2}{2 \zeta^2 \sigma_c^2}
\end{align}

Now consider the loss from the first term in \cref{apdx:decomposed_loss} due to a majority group. It must be at least
\begin{align*}
    \frac{K}{n} \left(1 - \sqrt{\frac{2}{d}}\zeta \inner{\vct{v}_s}{\vct{\beta}_{\ast}'} - \frac{\sqrt{g_{\maj}} R_s \|\vct{v}_c\|}{\sigma_c}\right)^2 &\leq \frac{g_{\maj} R_s^2}{2} \\
    1 - \sqrt{\frac{2}{d}}\zeta \inner{\vct{v}_s}{\vct{\beta}_{\ast}'} - \frac{\sqrt{g_{\maj}} R_s \|\vct{v}_c\|}{\sigma_c} &\leq \sqrt{\frac{ng_{\maj} R_s^2}{2K}} \\
    1 - \sqrt{g_{\maj}} R_s (\frac{\|\vct{v}_c\|}{\sigma_c} + \sqrt{\frac{n}{2K}}) &\leq  \sqrt{\frac{2}{d}}\zeta \inner{\vct{v}_s}{\vct{\beta}_{\ast}'}
\end{align*}
Note that $\sqrt{\frac{n}{2K}} \leq \sqrt{\frac{g_{maj}}{2}}$. Now if we have $R_s$ sufficiently smaller than $\frac{\sigma_c}{\sqrt{g_{maj}}\|\vct{v}_c\|}$ and $\frac{2}{g_{\maj}}$, we can guarantee that the RHS is at least some constant less than $1$, say $\frac{1}{\sqrt{2}}$. In this case, we have
\begin{align} \label{apdx:optimal_spurious_alignment}
    \inner{\vct{v}_{s}}{\vct{\beta}_{\ast}}^2 &\geq \frac{d}{4 \zeta^2}
\end{align}
Under these assumptions, it is clear from \cref{apdx:optimal_core_alignment} that we will also have
\begin{align}
    \frac{d}{4 \zeta^2} \gg \inner{\vct{v}_{c}}{\vct{\beta}_{\ast}}^2
\end{align}

Now we return to the original dataset, which contains minority groups and only a finite number of examples. Again, we have
\begin{align*}
    \vct{\beta}_{\ast} = (\mtx{\Psi}^{\top}\mtx{\Psi})^{\dagger}\mtx{\Psi}^{\top} \vct{y}
\end{align*}
Since we have removed the bias term, it is not hard to show that the matrix $\frac{1}{n} \mtx{\Psi}^{\top}\mtx{\Psi}$ has all eigenvalues of order $\Theta(\frac{1}{d})$. Now consider the norm of the difference $\|\frac{1}{n} \mtx{\Psi}^{\top}\mtx{\Psi} - \E_{\mathcal{D}_{\maj}}[\vct{\psi}_i \vct{\psi}_i^{\top}]\|$. With high probability, it will be of order $\OO(\frac{n_{\mino}}{nd} + \frac{1}{d}\sqrt{\frac{d}{n}} + \sqrt{\frac{\log{n}}{nd}}) = \OO(\frac{n^{-\gamma} + d^{-\Omega(\alpha)}}{d})$, where the first term corresponds to the inclusion of minority groups, the second term corresponds having a finite sample size, and the third term corresponds to using the true norm component. It follows that
\begin{align*}
    \left\|(\frac{1}{n} \mtx{\Psi}^{\top}\mtx{\Psi})^{\dagger} - (\E_{\mathcal{D}_{\maj}}[\vct{\psi}_i \vct{\psi}_i^{\top}])^{\dagger} \right\| &= \OO\left(d - \frac{d}{1 + \OO(n^{-\gamma} + d^{-\Omega(\alpha)}})\right) \\
    &= \OO(d(n^{-\gamma} + d^{-\Omega(\alpha)}))
\end{align*}
A similar argument shows that
\begin{align*}
    \|\mtx{\Psi}^\top \vct{y} - \E_{\mathcal{D}_\maj}[y_i \vct{\psi}_i]\| &= \OO(d^{-\frac{1}{2}} (n^{-\gamma} + d^{-\Omega(\alpha)}))
\end{align*}
Thus the change in alignment with a feature $\vct{v}$ is
\begin{align*} 
    \left\|\inner{\tilde{\vct{\beta}}_{\ast}}{\vct{v}} - \inner{\vct{\beta}_{\ast}}{\vct{v}}\right\| &= \left\| (\mtx{\Psi}^{\top}\mtx{\Psi})^{\dagger}\mtx{\Psi}^{\top} \vct{y} - \left(\E_{\mathcal{D}_{\maj}}[\vct{\psi}_i \vct{\psi}_i^{\top}]\right)^{\dagger} \E_{\mathcal{D}_{\maj}}[y_i \vct{\psi}_i] \right\| \|\vct{v}\| \\
    &\leq \Bigg\| \left((\mtx{\Psi}^{\top}\mtx{\Psi})^{\dagger} - \left(\E_{\mathcal{D}_{\maj}}[\vct{\psi}_i \vct{\psi}_i^{\top}]\right)^{\dagger}\right)\mtx{\Psi}^{\top} \vct{y} \\
    &\hspace{2.5cm}+ \left(\E_{\mathcal{D}_{\maj}}[\vct{\psi}_i \vct{\psi}_i^{\top}]\right)^{\dagger} (\mtx{\Psi}^{\top} \vct{y} - \E_{\mathcal{D}_{\maj}}[y_i \vct{\psi}_i]) \Bigg\| \|\vct{v}\| \\
    &\leq O\left(d(n^{-\gamma} + d^{-\Omega(\alpha)})(d^{-\frac{1}{2}}) + dd^{-\frac{1}{2}}(n^{-\gamma} + d^{-\Omega(\alpha)}) \right) \\
    \label{apdx:perturbation_error}
    &\leq \OO((n^{-\gamma} + d^{-\Omega(\alpha)}) \sqrt{d} )
\end{align*}

Replacing $g_{maj}$ with $2$,and combining equations \cref{apdx:optimal_core_alignment,apdx:optimal_spurious_alignment,apdx:perturbation_error} gives
\begin{align}
    |\inner{\vct{\beta}^{\ast}}{\vct{v}_{s}}| &\geq \frac{\sqrt{d}}{2 \zeta} \gg \sqrt{d}\left(\frac{R_s}{\zeta R_c} + \OO(n^{-\gamma} + d^{-\Omega(\alpha)}) \right) \geq |\inner{\vct{v}_{c}}{\vct{\beta}^{\ast}}|.
\end{align} 
Then by \cref{assumption:simplicity_bias_full}, we get
\begin{align}
    |f(\vb_s;\W_T,\z_{T})| \gg \frac{\sqrt{2}R_s}{R_c} + \OO(n^{-\gamma} + d^{-\Omega(\alpha)}) \geq |f(\vb_c;\W_T,\z_{T})|. \vspace{-2mm}
\end{align} 
which proves the theorem.

\begin{figure}
     \centering
    \begin{subfigure}[b]{0.24\textwidth}
         \centering
         \includegraphics[width=\textwidth]{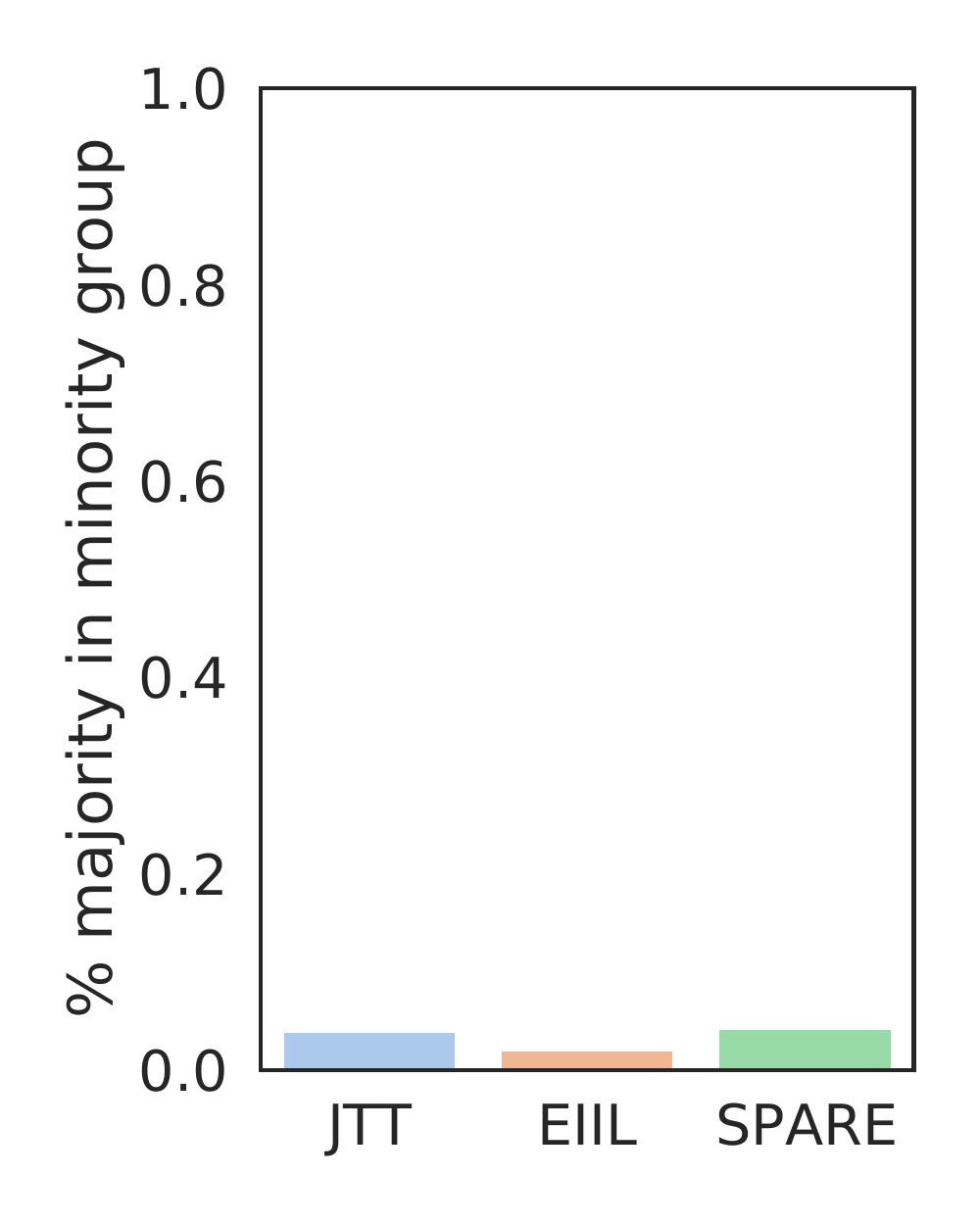}
         \caption{Waterbirds}
         \label{fig:waterbirds-major-in-minor}
     \end{subfigure}
     \begin{subfigure}[b]{0.24\textwidth}
         \centering
        \includegraphics[width=\textwidth]{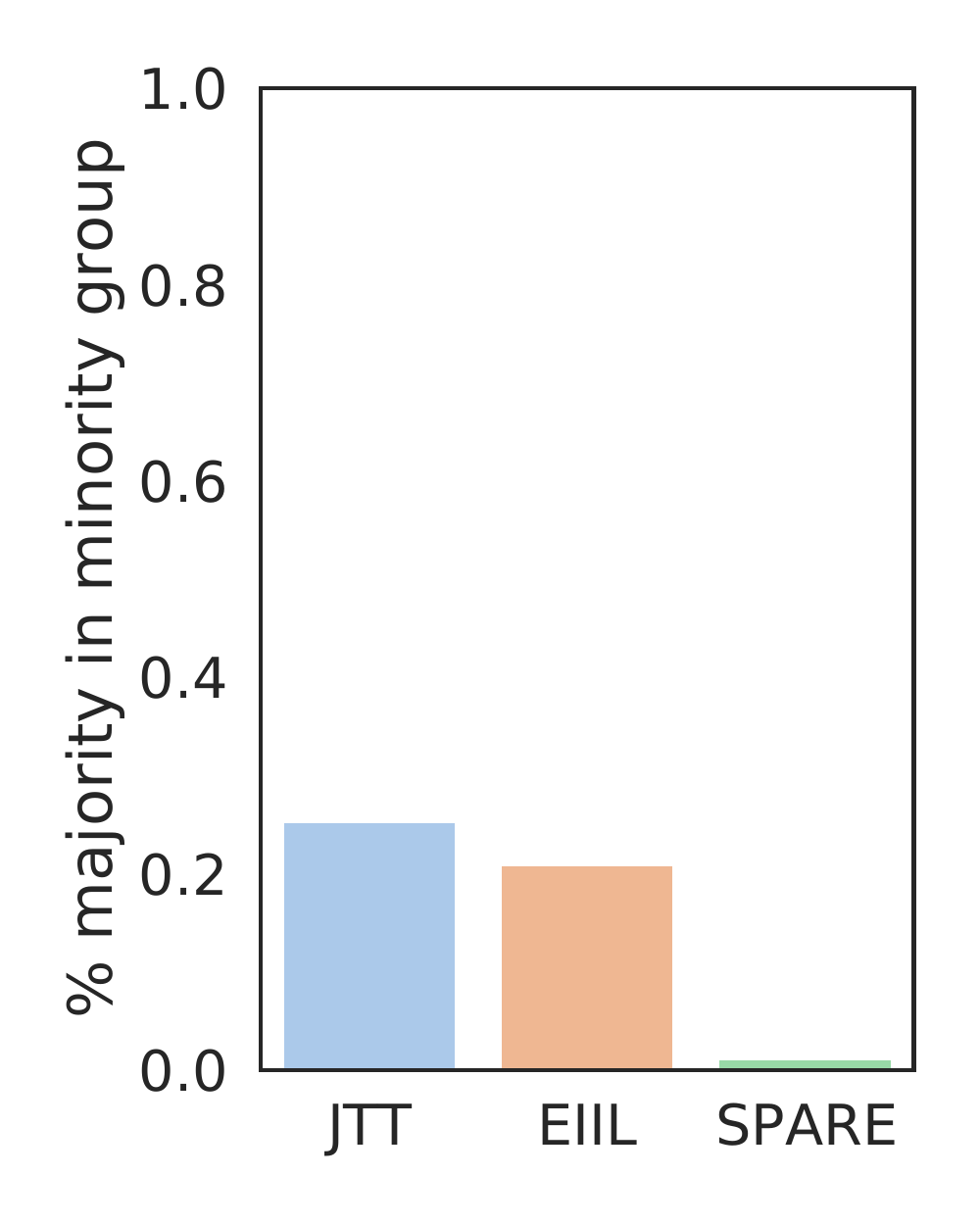}
         \caption{CelebA}
         \label{fig:celeba-major-in-minor}
     \end{subfigure}
     \begin{subfigure}[b]{0.24\textwidth}
         \centering
         \includegraphics[width=\textwidth]{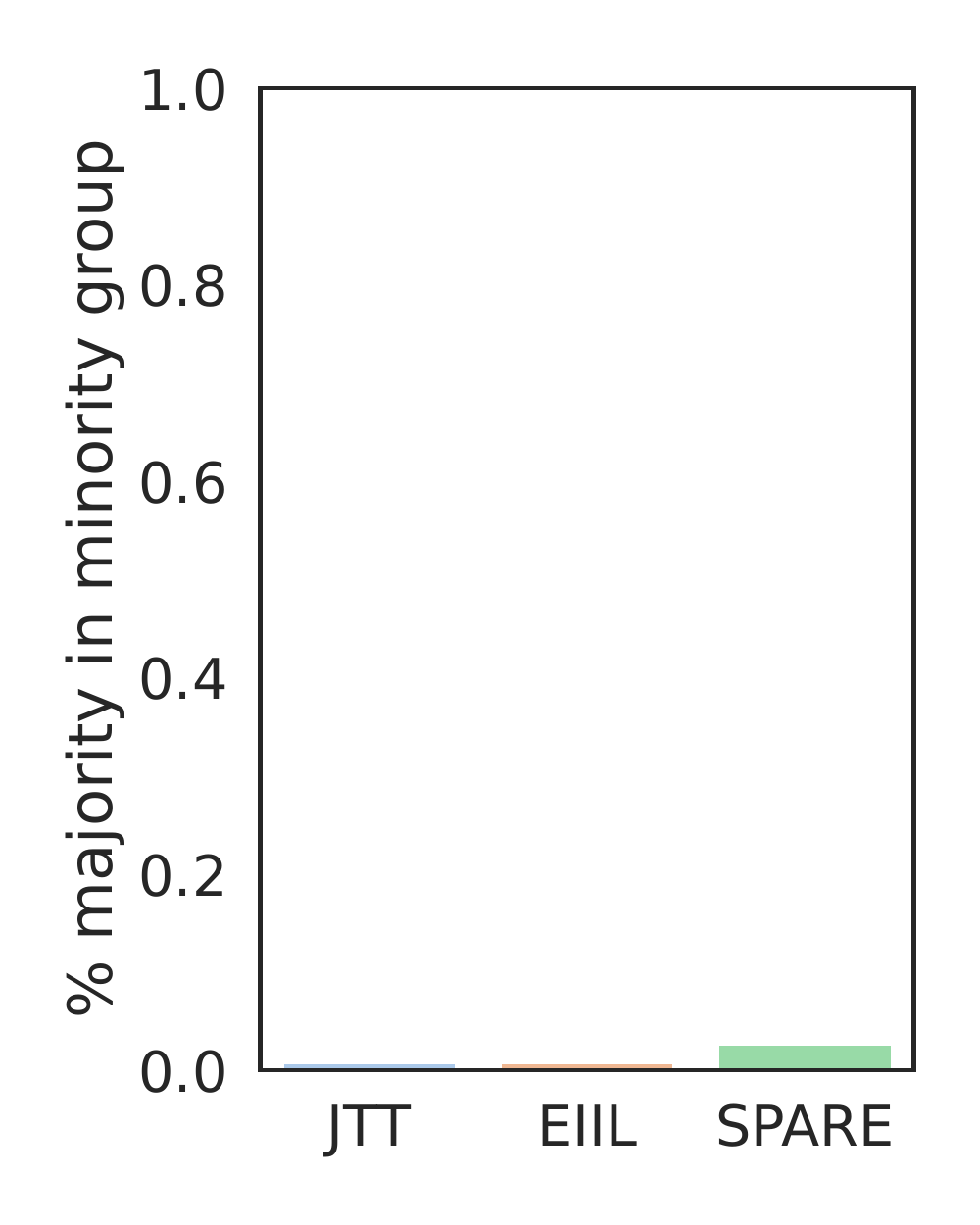}
         \caption{UrbanCars}
         \label{fig:urbancars-major-in-minor}
     \end{subfigure}
     \captionof{figure}{Fraction of majority examples inferred as minority. \alg not only identifies minority groups more accurately and correctly upweights them as evidenced in \cref{fig:minor-in-major}, but also does not identify a lot of majority examples as minority.}\looseness=-1\label{fig:major-in-minor}
\end{figure}

\section{MINIMAL MAJORITY IN MINORITY GROUPS BY \alg}

In \cref{sec:better}, we showed \alg's ability to accurately identify minority groups with minimal minority identified in the majority groups. We note that, as evidenced in \cref{fig:major-in-minor}, the presence of majority examples within the minority groups identified by \alg is also low. This reduced rate minimizes the likelihood of incorrect upweighting.

\begin{figure*}[t]
     \centering
    \begin{subfigure}[b]{0.32\textwidth}
         \centering
         \includegraphics[width=\textwidth]{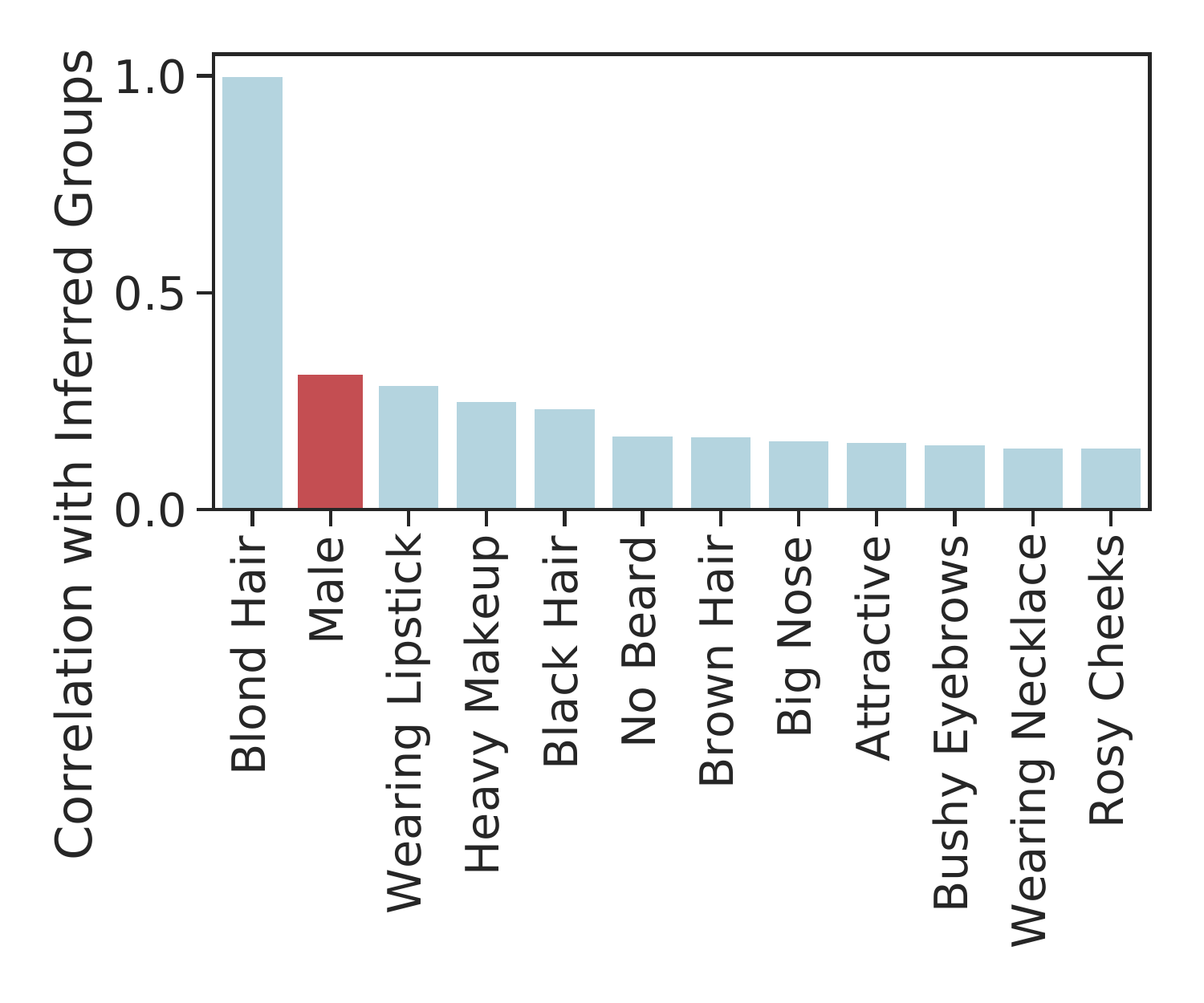}
         \caption{JTT}
         \label{fig:celeba-corr-jtt}
     \end{subfigure}
     \hfill
     \begin{subfigure}[b]{0.32\textwidth}
         \centering
        \includegraphics[width=\textwidth]{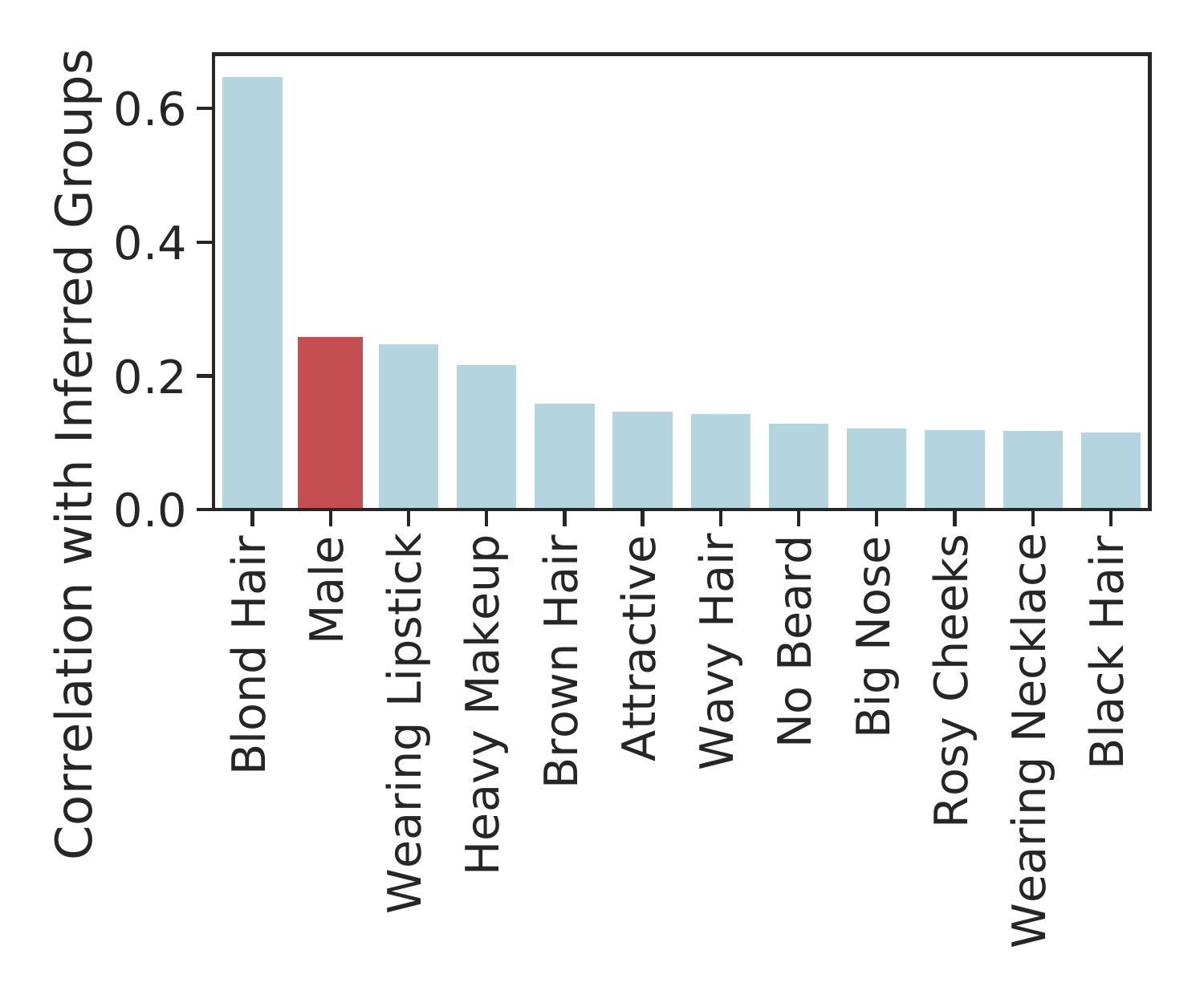}
         \caption{EIIL}
         \label{fig:celeba-corr-eiil}
     \end{subfigure}
    \hfill
     \begin{subfigure}[b]{0.32\textwidth}
         \centering
         \includegraphics[width=\textwidth]{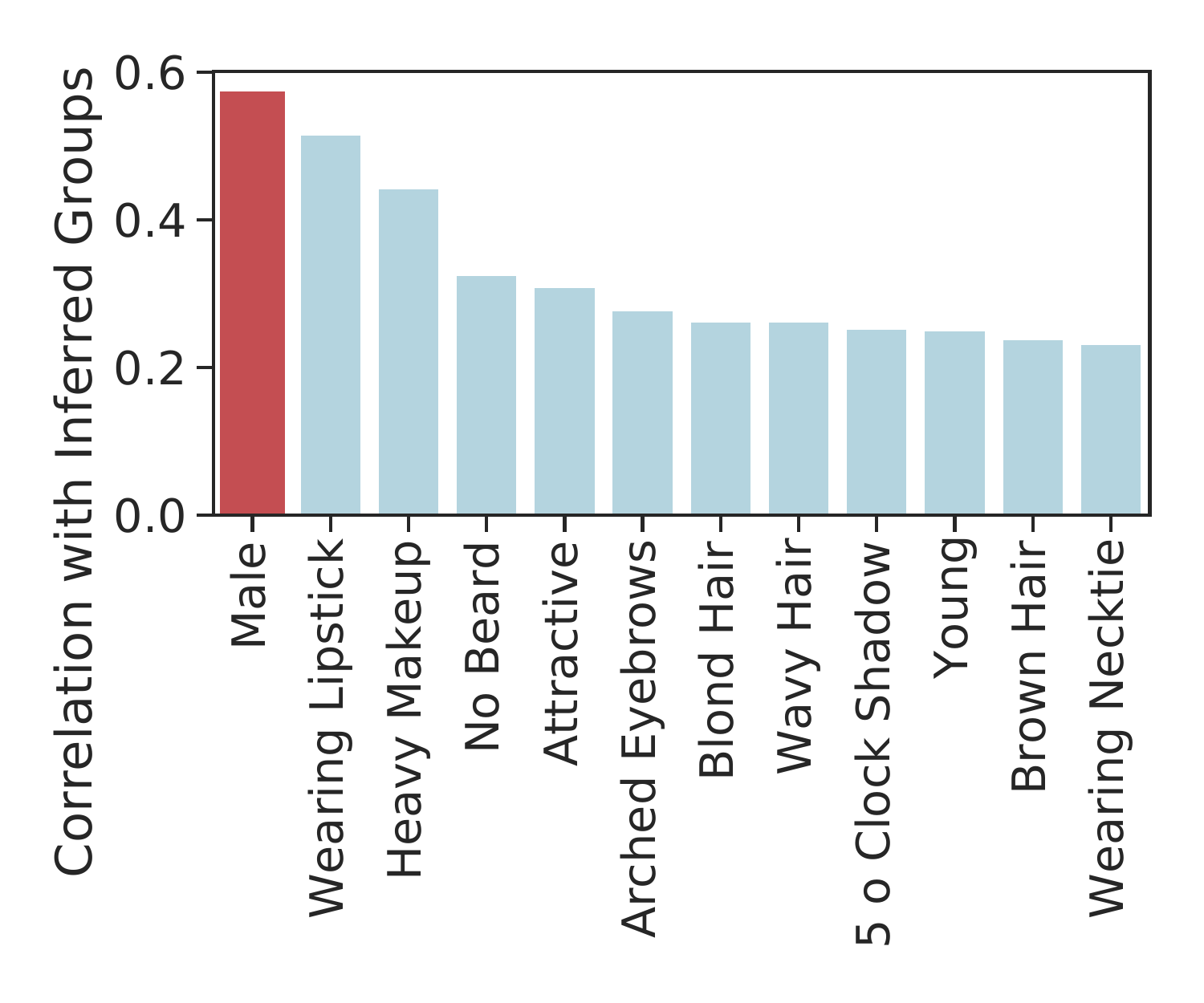}
         \caption{SPARE}
         \label{fig:celeba-corr-spare}
     \end{subfigure}
     \caption{Intercorrelation between the group inferred by state-of-the-art group inference methods (JTT, EIIL, \alg) and the attributes in the CelebA dataset, measured by Cramer's V \citep{cramer1999mathematical}. The higher the value, the more likely the inferred groups can be completely determined by the attribute. JTT and EIIL mainly separate majority and minority based on the class attribute (``Blond Hair'') while SPARE separates groups mainly based on the spurious attribute (``Male'', colored in red). \looseness=-1}
        \label{fig:celeba-corr}
\end{figure*}

\section{INTERCORRELATION OF INFERRED GROUPS AND ATTRIBUTES IN
CELEBA}

To understand the attributes used to separate the groups, we measure the intercorrelation using Cramer's V \citep{cramer1999mathematical} between groups inferred by different group inference methods (JTT, EIIL, and \alg) and the attributes in the CelebA dataset. The metric allows us to measure the likelihood that a given attribute can completely determine the inferred groups. 

As illustrated in \cref{fig:celeba-corr}, JTT and EIIL show a higher Cramer's V value for the class attribute ``Blond Hair'', indicating that their inferred groups are mainly based on this attribute. On the other hand, \alg (SPARE) exhibits a higher value for the spurious attribute ``Male'', colored in red, demonstrating that it more effectively separates groups based on the spurious attribute.

\section{EXPERIMENTATION DETAILS}\label{sec:experiment}

\subsection{Datasets}\label{sec:cmnist}
\paragraph{CMNIST} We created a colored MNIST dataset with spurious correlations by using colors as spurious attributes following the settings in \cite{zhang2022correct}. First, we defined an image classification task with 5 classes by grouping consecutive digits (0 and 1, 2 and 3, 4 and 5, 6 and 7, 8 and 9) into the same class. From the train split, we randomly selected 50,000 examples as the training set, while the remaining 10,000 samples were used as the validation set. The test split follows the official test split of MNIST.

For each class $y_i$, we assigned a color $\vb_s$ from a set of colors $\mathcal{A}$=\{\texttt{\#ff0000}, \texttt{\#85ff00}, \texttt{\#00fff3}, \texttt{\#6e00ff}, \texttt{\#ff0018}\} as the spurious attribute that highly correlates with this class, represented by their hex codes, to the foreground of a fraction $p_{corr}$ of the training examples. This fraction represents the majority group for class $y_i$. The stronger the spurious correlation between class $y_i$ and the spurious attribute $\vb_{s}$, the higher the value of $p_{corr}$. The remaining $1-p_{corr}$ training examples were randomly colored using a color selected from $\mathcal{A}\setminus \vb_{s}$. 
In our experiments, we set $p_{corr}=0.995$ to establish significant spurious correlations within the dataset. 

\paragraph{Waterbirds} is introduced by \cite{sagawa2019distributionally} to study the spurious correlation between the background (land/water) and the foreground (landbird/waterbird) in image recognition. 
Species in Caltech-UCSD Birds-200-2011 (CUB-200-2011) dataset \citep{WahCUB_200_2011} are grouped into two classes, waterbirds and landbirds. All birds are then cut and pasted onto new background images, with waterbirds more likely to appear on water and landbirds having a higher probability on land. 
There are 4795 training examples in total, 3498 for landbirds with land background, 184 for landbirds with water background, 56 for waterbirds with land background, and 1057 for waterbirds with water background. 

\paragraph{CelebA} is a large-scale face attribute dataset comprised of photos of celebrities. Each image is annotated with 40 binary attributes, in which ``blond hair'' and ``male'' are commonly used for studying spurious correlations. Specifically, gender is considered a spurious feature for hair color classification. The smallest group is blond male. 

\paragraph{UrbanCars} The UrbanCars dataset is introduced by \cite{li2023whac} to explore the impact of multiple spurious correlations in image classification. Each image features a car centrally placed against a natural scene background, accompanied by a co-occurring object to the right. The goal is to classify the car's body type while accounting for two spurious attributes: the background (BG) and the co-occurring object (CoObj), which are correlated with the target label. The labels share a binary space, consisting of two classes: urban and country. The dataset is partitioned into 8 groups based on various combinations of these labels. The training set manifests strong spurious correlations of 0.95 in strength for both BG and CoObj.

UrbanCars is assembled from multiple source datasets, including Stanford Cars \citep{krause20133d} for the car objects and labels, Places \citep{zhou2018places} for the backgrounds, and LVIS \citep{gupta2019lvis} for the co-occurring objects. Backgrounds and co-occurring objects are selected to fit the ``urban'' and ``country'' classes.

For evaluation, the authors employ a range of metrics, including In Distribution Accuracy (I.D. Acc), BG Gap, CoObj Gap, and BG+CoObj Gap. These metrics gauge both the model's overall performance and its robustness in handling group shifts due to individual or combined spurious attributes.

\subsection{Hyperparameters}
\label{sec:hyper}
{We used SGD as the optimization algorithm to maintain consistency with the existing literature.} The hyperparameters employed in our experiments on spurious benchmarks are detailed in \cref{tab:hyper}. For the Waterbirds, CelebA and UrbanCars datasets, we tuned the learning rate within the range of \{1e-4, 1e-5\} and weight decay within the range of \{1e-1, 1e-0\}. These ranges were determined based on the ranges of optimal hyperparameters used by the current state-of-the-art algorithms \citep{creager2021environment,liu2021just,sagawa2019distributionally,nam2021spread,zhang2022correct}. The batch sizes and total training epochs remained consistent with those used in these prior studies. To determine the epoch for separating groups, we performed clustering on the validation set while training the model on the training set to maximize the minimum recall of \alg's clusters with the groups in the validation set. As mentioned in \cref{sec:method}, we decided the number of clusters and adjusted the sampling power for each class based on Silhouette scores. Specifically, when the Silhouette score was below 0.9, a sampling power of 2 or 3 was applied, while a sampling power of 1 was used otherwise. It is important to note that other algorithms tuned hyperparameters, such as epochs to separate groups and upweighting factors, by maximizing the worst-group accuracy of fully trained models on the validation set, which is more computationally demanding than the hyperparameter tuning of \alg.  

\begin{table}[h]
\caption{Hyperparameters used for the reported results on different datasets.}
\label{tab:hyper}
\vskip 0.15in
\begin{center}
\begin{small}
\begin{sc}
\begin{tabular}{lcccc}
\toprule
Dataset & CMNIST & Waterbirds & CelebA & UrbanCars \\
\midrule
Learning rate & 1e-3 & 1e-4 & 1e-5 & 1e-4 \\
Weight decay & 1e-3 & 1e-1 & 1e-0 & 1e-1 \\
Batch size & 32 & 128 & 128 & 128 \\
Training epochs & 20 & 300 & 50 & 300 \\
\midrule
Group separation epoch & 2 & 2 & 1 & 2 \\
Silhouette scores & [0.997,0.978,0.996,0.991,0.996] & [0.886,0.758] & [0.924,0.757] & [0.849,0.872] \\
Sampling power& [1,1,1,1,1]  & [3,3]  & [1,2] & [2,2]\\
\bottomrule
\end{tabular}
\end{sc}
\end{small}
\end{center}
\vskip -0.1in
\end{table}

\subsection{Choices of Model Outputs}
In our experiments, we found the worst-group accuracy gets the most improvement when \alg uses the outputs of the last linear layer to separate the majority from the minority for CMNIST, Waterbirds and UrbanCars and use the second to last layer (i.e., the feature embeddings inputted to the last linear layer) to identify groups in CelebA. We speculate that this phenomenon can be attributed to the increased complexity of the CelebA dataset compared to the other two datasets, as employing a higher output dimension help identify groups more effectively.

\subsection{Dependency on the Clustering Algorithm}
\label{sec:cluster-alg}

The performance of \alg is not sensitive to the clustering algorithm. The key to \alg is \textbf{clustering the entire model output early in training}. While $k$-means easily scales to medium-sized datasets, $k$-median is more suitable for very large datasets, as it can be formulated as a submodular maximization problem \citep{wolsey1982analysis} for which fast and scalable distributed \citep{mirzasoleiman2013distributed,mirzasoleiman2015lazier} and streaming \citep{badanidiyuru2014streaming} algorithms are available.

\subsection{Clustering Details}
Clustering was performed on all data samples within the same class. It's important to note that k-means doesn't require loading all the data into memory and operates in a streaming manner. As an alternative, we also discussed the possibility of using the k-medoids clustering algorithm and its distributed implementation which uses submodular optimization and easily scales to millions of examples in \cref{sec:method}. 
In \cref{tab:cluster}, we present the wall-clock times for k-means clustering on CMNIST, Waterbirds, CelebA, UrbanCars and Restricted ImageNet. It shows that the cost of clustering is negligible when compared to the cost of training.

\begin{table}[t]
\caption{Wall-clock times for k-means clustering on Waterbirds, CelebA, CMNIST, and Restricted ImageNet datasets. }
\label{tab:cluster}
\begin{center}
\begin{tabular}{rrrrr}
\toprule
\multicolumn{1}{c}{\bf CMNIST}  
&\multicolumn{1}{c}{\bf Waterbirds} &\multicolumn{1}{c}{\bf Celeba} &\multicolumn{1}{c}{\bf UrbanCars} &\multicolumn{1}{c}{\bf Restricted ImageNet}
\\ \midrule
0.46s & 0.07s & 31.8s & 0.57s & 2s \\
\bottomrule
\end{tabular}
\end{center}
\end{table}

\begin{table}[t]
\caption{Wall-clock runtime comparison of  \alg and SOTA 2-stage algorithms. }
\label{tab:wallclock}
\begin{center}
\begin{tabular}{rrrrr}
\toprule
\multicolumn{1}{c}{\bf ERM}  &\multicolumn{1}{c}{\bf JTT}
&\multicolumn{1}{c}{\bf CnC}  &\multicolumn{1}{c}{\bf SSA}
&\multicolumn{1}{c}{\bf \alg}
\\ \midrule
1h12m & 9h5m & 4h25m & 2h15m & 1h16m \\
\bottomrule
\end{tabular}
\end{center}
\end{table}

\subsection{Training Cost}
\label{sec:train-cost}
 \cref{tab:wallclock} shows a all-clock runtime comparison of \alg and SOTA 2-stage algorithms on Waterbirds. JTT initially trains the identification model for a specific number of epochs and then upsamples misclassified examples by a substantial factor to train the robust model. As a result, the training cost is influenced not just by the training of the identification model but also by the considerable volume of upsampled training data used in the robust model's training. For instance, in the case of CelebA, JTT trains the identification model for just one epoch but then upsamples all misclassified examples (approximately 1/10 of the training set) by a factor of 50. This leads to a training set roughly six times the original size. In this scenario, the large volume of upsampled training data significantly increases the training cost, while the training time for the identification model is almost negligible.

\begin{figure*}[t]
\begin{center}
\includegraphics[width=.59\textwidth]{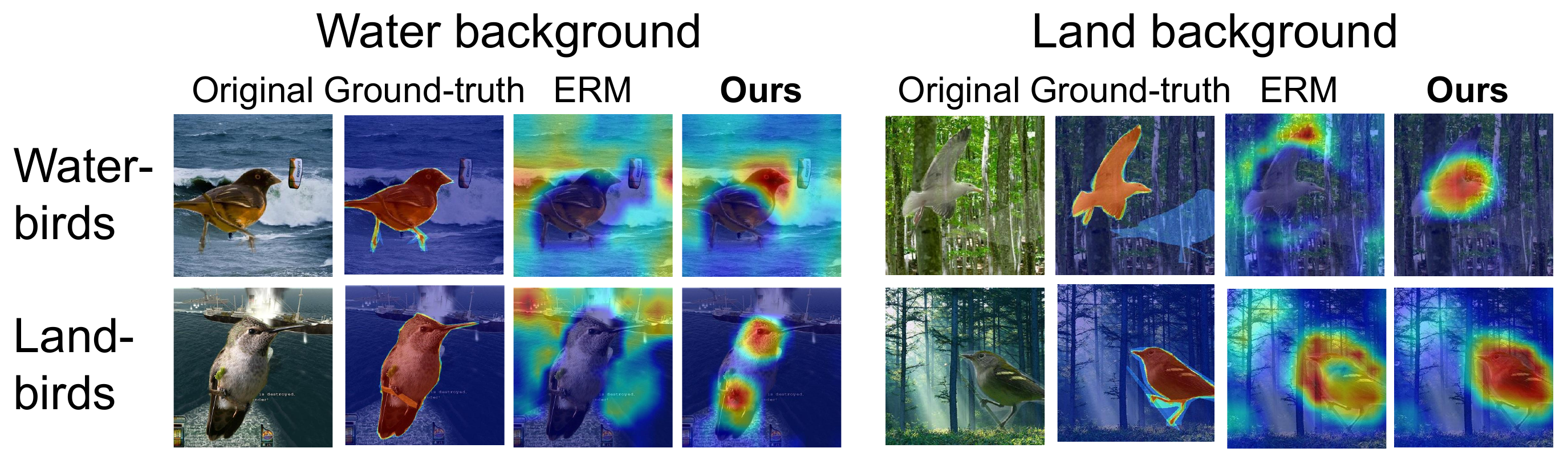}
\includegraphics[width=.38\textwidth]{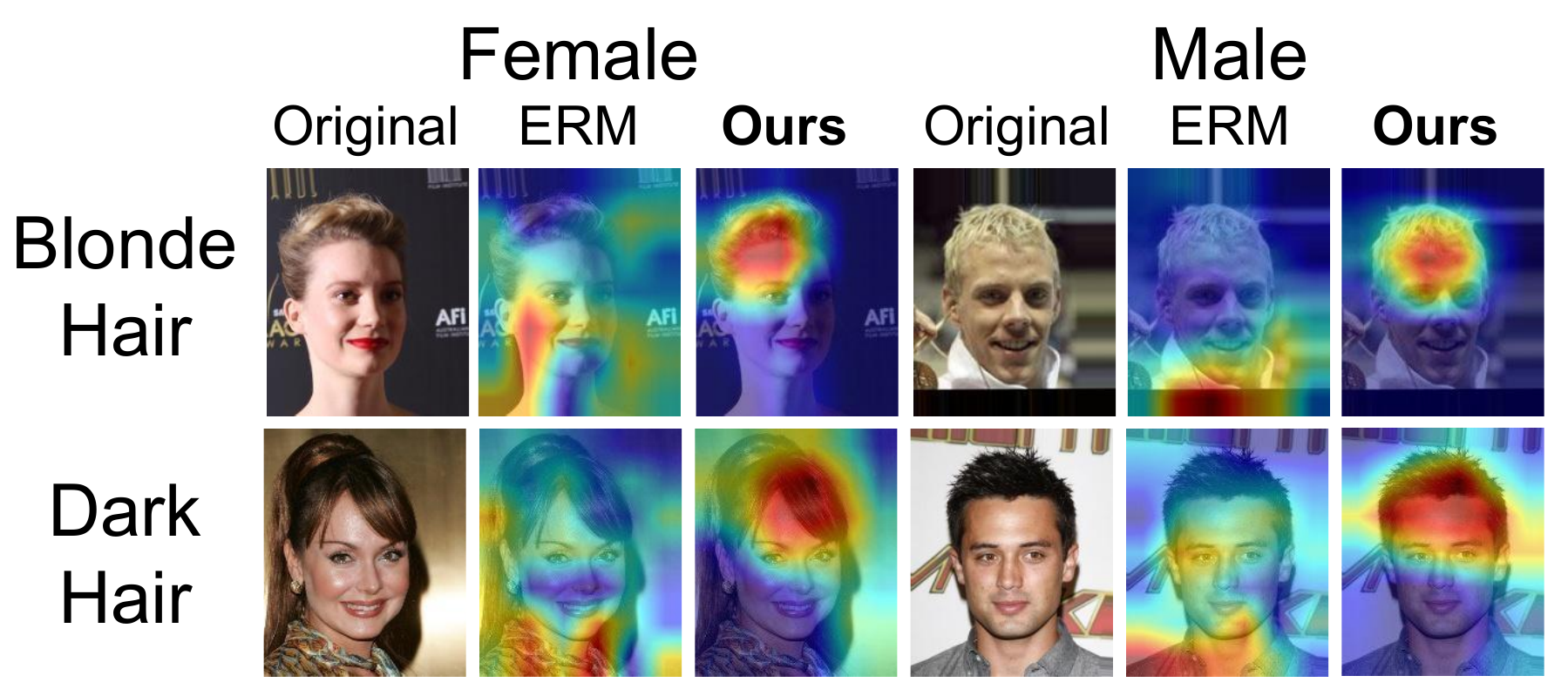}
\end{center}
\caption{GradCAM Visualization. Warmer colors correspond to the pixels that are weighed more in making the final classification. %
\alg allows learning the core features instead of spurious ones. %
}\label{fig:gradcam}
\end{figure*}

\newpage
\section{GRADCAM VISUALIZATIONS: \alg HELPS THE LEARNING OF CORE
FEATURES}
\cref{fig:gradcam} compares GradCAM \citep{selvaraju2017grad} visualizations depicting saliency maps for samples from Waterbirds with water and land backgrounds (left), and from CelebA with different genders (right), when ResNet50 is trained by ERM vs. \alg. Warmer colors %
indicate the pixels that the model considered more important for making the final classification, based on
gradient activations. We see that training with \alg allows the model to learn the core feature, instead of the spurious features.\looseness=-1

\section{DISCOVERING SPURIOUS FEATURES}
\label{sec:imagenet-app}

\subsection{Restricted ImageNet}
We use Restricted ImageNet proposed in \cite{tsipras2018robustness} which contains 9 superclasses of ImageNet. The classes and the corresponding ImageNet class ranges are shown in \cref{tab:restrict-imagenet}.

\begin{table}[h]
\caption{Classes included in Restricted ImageNet and their corresponding ImageNet class ranges.}
\label{tab:restrict-imagenet}
\centering
\begin{tabular}{c|c}
\toprule
 Restricted ImageNet Class & ImageNet class range \\
 \midrule
 dog & 151-268 \\
 cat & 281-285 \\
 frog & 30-32 \\
 turtle & 33-37 \\
 bird & 80-100 \\
 primate & 365-382 \\
 fish & 389-397 \\
 crab & 118-121 \\ 
 insect & 300-319 \\
\bottomrule
    \end{tabular}
    
\end{table}

\subsection{Experimental Settings}
When training on Restricted ImageNet, we use ResNet50 \citep{he2016deep} from the PyTorch library \citep{pytorch} with randomly initialized weights instead of pretrained weights. We followed the hyperparameters specified in \cite{goyal2017accurate}: the model was trained for 90 epochs, with an initial learning rate of 0.1. The learning rate was reduced by a factor of 0.1 at the 30th, 60th, and 80th epochs. During training, we employed Nesterov momentum of 0.9 and applied a weight decay of 0.0001.

\subsection{Investigation on Groups Identified by EIIL vs. \alg}

\paragraph{Evaluation Setup. } As no group-labeled validation set is available to tune the epoch in which the groups are separated, we tried separating groups using ERM models trained for various numbers of epochs. %
Since both EIIL and \alg identify the groups early (EIIL infers groups on models trained with ERM for 1 epoch for both Waterbirds and CelebA, and 5 epochs for CMNIST; the group separation epochs for \alg are epoch 1 or 2 for the three datasets, as shown in \cref{tab:hyper}), we tuned the epoch to separate groups in the range of \{2,4,6,8\} for both algorithms. This tuning was based on the average test accuracy achieved by the final model, as the worst-group accuracy is undefined without group labels. 
Interestingly, while \alg did not show sensitivity to the initial epochs on Restricted ImageNet, EIIL achieved the highest average test accuracy when the initial models were trained for 4 epochs using ERM. We manually labeled examples with their groups for test data.

\textbf{EIIL finds groups of misclassified examples while \alg finds groups with spurious features.} We observed that EIIL effectively separates examples that have 0\% classification accuracy as the minority group, as demonstrated in \cref{tab:eiil-acc}. This separation is analogous to the error-splitting strategy employed by JTT \citep{liu2021just} when applied to the same initial model. 
This similarity in behavior is also discussed in \citep{creager2021environment}. 
Instead of focusing on misclassified examples, \alg separates the examples that are learned early in training. \cref{tab:spare-acc} shows that the first cluster found by \alg have almost 100\% accuracy, indicating that the spurious feature is learned for such examples. Downweighting examples that are learned early allows for effectively mitigating the spurious correlation. 

\textbf{\alg upweights outliers less than EIIL. } Heavily upweighting misclassified examples can be problematic for this more realistic dataset than the spurious benchmarks as the misclassified ones are likely to be outliers, noisy-labeled or contain non-generalizable information. \cref{tab:eiil-acc} shows that groups inferred by EIIL are more imbalanced, which makes EIIL upweights misclassified examples more than \alg. As shown in \cref{tab:imagenet}, this heavier upweighting of misclassified examples with EIIL drops accuracy not only for the minority groups but also for the overall accuracy. 
Therefore, we anticipate that this effect would persist or become even more pronounced for methods like JTT, which directly identify misclassified examples as the minority group. In contrast, \alg separates groups based on the spurious feature that is learned early, and upweights the misclassified examples less than other methods due to the more balanced size of the clusters. This allows \alg to more effectively mitigate spurious correlations than others.

\begin{table}[t]
\caption{Accuracy (\%) of training examples in different classes of Restricted ImageNet in the two environments inferred by EIIL. EIIL trains models with Group DRO on the inferred environments, resulting in up-weighting misclassified examples in Env 2.}
    \label{tab:eiil-acc}
    \centering
    \begin{tabular}{c|ccccccccc}
    \toprule
 Class & dog & cat & frog & turtle & bird & primate & fish & crab & insect \\
 \midrule
 Env 1 ERM acc & 98 & 37 & 26 & 62 & 76 & 78 & 78 & 71 & 90 \\
 Env 2 ERM acc & 0 & 0 & 0 & 0 & 0 & 0 & 0 & 0 & 0 \\
 \midrule
  Env 1 size & 144378 & 488 & 457 & 2875 & 17157 & 11233 & 6817 & 2172 & 21112 \\
 Env 2 size & 3495 & 6012 & 3443 & 3625 & 9984 & 12167 & 4417 & 3028 & 4888 \\
\bottomrule
    \end{tabular}
    
\end{table}

\begin{table}[!ht]
\caption{Accuracy (\%) of training examples in different classes of Restricted ImageNet in the two groups inferred by \alg at epoch 8.}
    \label{tab:spare-acc}
    \centering
    \begin{tabular}{c|ccccccccc}
    \toprule
 Class & dog & cat & frog & turtle & bird & primate & fish & crab & insect \\
 \midrule
 Cluster 1 ERM acc & 100 & 100 & 100 & 100 & 100 & 99 & 100 & 100 & 100 \\
 Cluster 2 ERM acc & 64 & 9 & 11 & 14 & 28 & 13 & 27 & 16 & 36 \\
 \midrule
  Cluster 1 size & 130541 & 3236 & 1578 & 2684 & 18870 & 12158 & 7331 & 2566 & 18974 \\
 Cluster 2 size & 17332 & 3264 & 2322 & 3816 & 8271 & 11242 & 3903 & 2634 & 7026 \\
\bottomrule
    \end{tabular}
\end{table}

\section{REPRODUCIBILITY}
Each experiment was conducted on one of the following GPUs: NVIDIA A40 with 45G memory, NVIDIA RTX A6000 with 48G memory, and NVIDIA RTX A5000 with 24G memory.